\documentclass{article} %

\usepackage{iclr2025_conference,times}

\usepackage{hyperref}
\usepackage{url}

\usepackage{graphicx}
\usepackage{subfigure}
\usepackage{enumitem}

\usepackage{bbding}
\usepackage{amsfonts}
\usepackage{amssymb}

\usepackage{threeparttable}
\usepackage{multicol,multirow}
\usepackage{booktabs} %
\usepackage{amsthm}
\usepackage{amsmath}
\usepackage{multirow}
\usepackage{makecell}

\def\red{\textcolor{red}}

\usepackage{afterpage}

\usepackage{xspace}
\newcommand{\methodname}{TCR\xspace}

\usepackage{colortbl}  %
\usepackage{xcolor}
\usepackage{array}   %

\usepackage[ruled,linesnumbered]{algorithm2e}

\usepackage{hyperref}
\hypersetup{
    colorlinks=true,
    linkcolor=red,
    citecolor=teal,
    urlcolor=cyan,
}
\usepackage{pifont}
\let\oldding\ding%
\renewcommand{\ding}[2][1]{\scalebox{#1}{\oldding{#2}}}%
\usepackage{wrapfig}

\usepackage{etoc}
\etocdepthtag.toc{mtchapter}
\etocsettagdepth{mtchapter}{subsubsection}
\etocsettagdepth{mtappendix}{none}

\def\mytitle{Test-time Adaptation for Cross-modal \\Retrieval with Query Shift}

\title{\mytitle}
\author{%
  Haobin Li$^1$\quad
  Peng Hu$^1$\quad
  Qianjun Zhang$^2$\quad
  Xi Peng$^{1,3}$\quad 
  XitingLiu$^4$\quad
  Mouxing Yang$^1$\thanks{Corresponding author.}\\
  College of Computer Science, Sichuan University, China.$^1$\quad
  Southwest Jiaotong University, China.$^2$\\
  State Key Laboratory of Hydraulics and Mountain River Engineering, Sichuan University, China.$^3$\\
  Georgia Institute of Technology, USA.$^4$\\
  \{haobinli.gm, penghu.ml, pengx.gm, liu.xt0617, yangmouxing\}@gmail.com, zqjblue@163.com
}

\iclrfinalcopy %
\begin{document}

\maketitle

\begin{abstract}
The success of most existing cross-modal retrieval methods heavily relies on the assumption that the given queries follow the same distribution of the source domain. 
However, such an assumption is easily violated in real-world scenarios due to the complexity and diversity of queries, thus leading to the query shift problem.
Specifically, query shift refers to the online query stream originating from the domain that follows a different distribution with the source one.
In this paper, we observe that query shift would not only diminish the uniformity (namely, within-modality scatter) of the query modality but also amplify the gap between query and gallery modalities. 
Based on the observations, we propose a novel method dubbed Test-time adaptation for Cross-modal Retrieval (TCR). 
In brief, TCR employs a novel module to refine the query predictions (namely, retrieval results of the query) and a joint objective to prevent query shift from disturbing the common space, thus achieving online adaptation for the cross-modal retrieval models with query shift.
Expensive experiments demonstrate the effectiveness of the proposed TCR against query shift. 
The code will be released upon acceptance.
\end{abstract}

\section{Introduction}
\label{sec:introduction}
\begin{figure*}[t]
\centering
\includegraphics[width=0.98\linewidth]{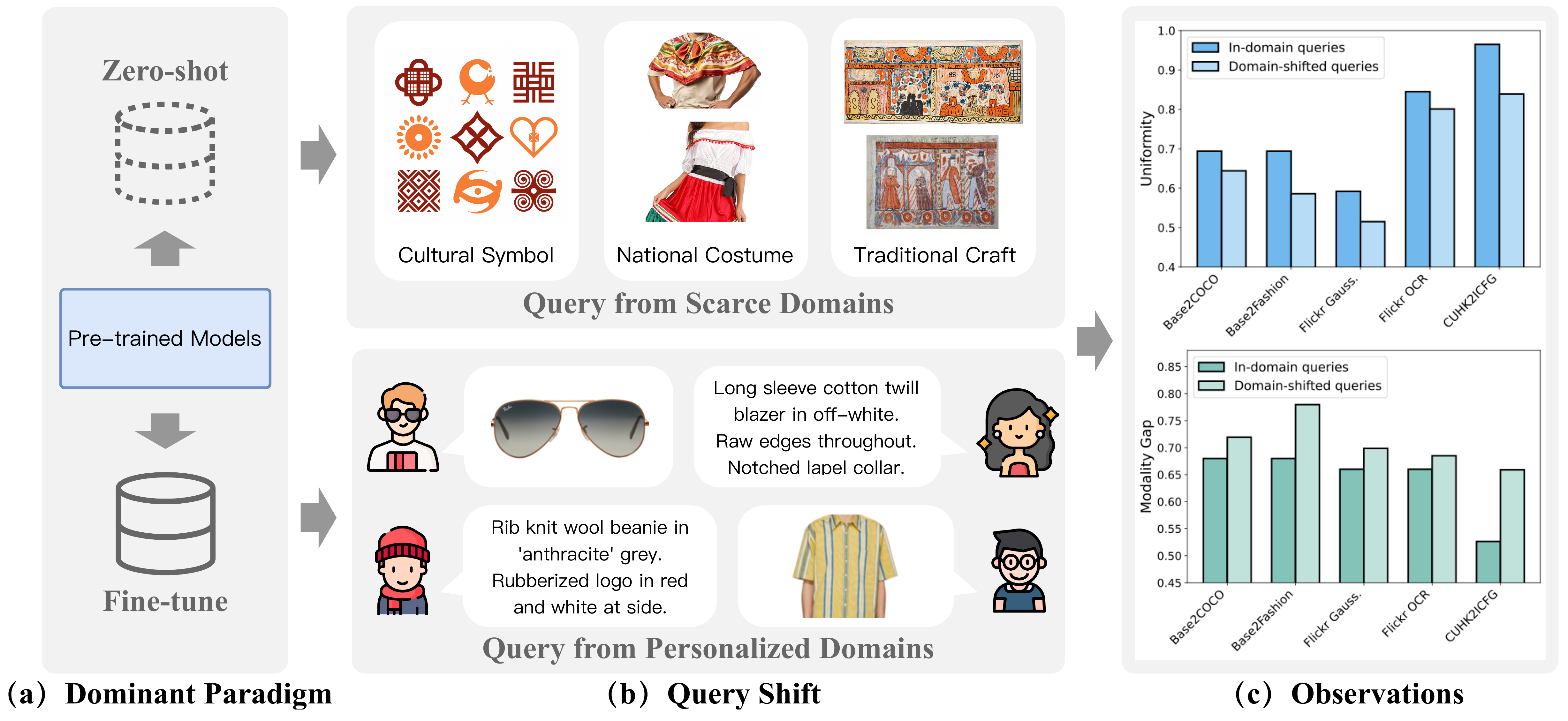}
\vspace{-0.05in}
\caption{
(a) \textbf{Dominant Paradigm}: 
the pre-trained models embrace powerful zero-shot retrieval capacity and could be fine-tuned on domain-specific data for customization, which has emerged as the dominant paradigm for cross-modal retrieval.
(b) \textbf{Query Shift}: 
the performance of the paradigm would be significantly degraded when encountering the query shift problem.
On the one hand, collecting sufficient data to tailor the pre-trained models for scarce domains is daunting and even impossible. 
On the other hand, as the saying goes, ``Different strokes for different folks", even fine-tuned models cannot accommodate all personalized domains.
(c) \textbf{Observations}: we study the query shift problem for cross-modal retrieval and reveal the following observations. 
Namely, query shift not only diminishes the uniformity of the query modality but also amplifies the modality gap between the query and gallery modalities, undermining the well-structured common space inherited from pre-trained models.
}
\vspace{-0.2in}
\label{fig:fig1}
\end{figure*}

Given queries of interest, cross-modal retrieval~\citep{VSE++,SCAN,CLIP-REID,Fashionklip} try to associate some relevant samples from the gallery set across various modalities, supporting numerous applications such as intelligent surveillance and search engine.
The key to cross-modal retrieval is learning a well-established common space, hoping to distinguish different instances within the same modality while gathering the same instance across different modalities.
Recently, the pre-trained models~\citep{ALIGN,TCL,BLIPV2,Huang} have emerged as the dominant paradigm for cross-modal retrieval.
As shown in Fig.~\ref{fig:fig1}(a), after acquiring generic knowledge from the source domain, the pre-trained models can either perform zero-shot retrieval in the target domains or be fine-tuned on domain-specific data for customization. 

Despite the promising performance of the pre-trained models, their success heavily relies on the assumption that the given queries exactly follow the same distribution from the source domain, which is hard to satisfy in real-world applications.
Specifically, as shown in Fig.~\ref{fig:fig1}(b), inquirers might embrace different cultural backgrounds or enjoy their individual preferences, resulting in the online query stream derived from either scarce or highly personalized domains.
Clearly, such out-of-domain queries violate the identical distribution assumption and thus lead to the \textit{query shift} problem.
As a result, the existing cross-modal retrieval models fail to handle the query shift and inevitably suffer from significant performance degradation, leaving an urgent need to develop an online adaptation method for addressing the query shift problem.

As one of the most effective paradigms in reconciling distribution shifts, Test-Time Adaptation (TTA) methods~\citep{Tent,Rdumb,FOA} work by continually updating the given source model using the online target data stream.
Although achieving great success, it is intractable to adopt existing TTA methods for cross-modal retrieval with query shift due to the following two reasons.
On the one hand, most existing TTA methods focus on the unimodal setting while overlooking the complexity of the query shift in the cross-modal setting.
More specifically, the query shift in the cross-modal setting would not only affect the intra-modality distribution but also hinder the cross-modal alignment.
On the other hand, most existing TTA methods are specifically designed for the recognition task, which would struggle with the heavy noise from the query predictions if simply applied to the retrieval task.
Intuitively, for a given sample or query, it has a random probability of $1/K$ or $1/N$ to be correctly associated with the desirable category or cross-modal counterpart in the recognition task or retrieval task, where $K$ and $N$ denote the class number and candidate number, respectively, with $N\gg K$.
%

To specifically develop a TTA method for cross-modal retrieval with the query shift, we first present two key observations as illustrated in Fig.~\ref{fig:fig1}(c). 
To summarize, we conclude that the query shift would diminish the uniformity of the query modality, prohibiting discrimination between diverse queries in the common space.
Moreover, query shift would amplify the modality gap between query and gallery modalities, undermining the well-constructed common space established by the pre-trained models.

Based on the above observations, we propose achieving robust cross-modal retrieval against the query shift by endowing the existing TTA methods with the capacity to manipulate both the modality uniformity and modality gap. 
To be specific, we propose a novel method, dubbed Test-time adaptation for Cross-modal Retrieval (TCR), which consists of a novel query prediction refinement module and a novel joint objective function.
First, the query prediction refinement module is adopted to refine the retrieval results of the existing TTA methods and thus obtain the retrieval-favorable predictions for queries.
After that, the joint objective is employed on the refined query predictions to achieve online adaptation for cross-modal retrieval models under query shift.
More specifically, the joint objective function is composed of three individual losses that embrace the following merits.
To enhance the uniformity of the query modality, the intra-modality uniformity learning loss performs contrast between queries and their respective centers, thus guaranteeing the discrimination between queries.
To rectify the modality gap between the query and gallery modalities, the inter-modality gap learning loss narrows the difference between the query and gallery modalities with the plausible constraint estimated from off-the-shelf models, thus inheriting the well-established common space.
To prevent overfitting on noisy query predictions, the noise-robust adaptation loss amplifies the contribution of high-confident predictions while alleviating the noisy ones with a self-adaptive threshold.   
The major contributions of this work could be summarized as follows.
\begin{itemize}
    \item To the best of our knowledge, this work could be one of the first studies on the query shift problem, revealing its underlying impacts on cross-modal retrieval. Specifically, the query shift would not only diminish the uniformity of the query modality but also amplify the modality gap between query and gallery modalities, undermining the well-established common space derived from the source model.
    \item We propose a novel test-time adaptation method named TCR. TCR first employs a novel module to refine the query predictions, thus supporting the existing TTA methods for cross-modal retrieval. Then, TCR adopts a novel objective function that can not only manipulate both the modality uniformity and modality gap but also prevent the model from overfitting noisy query predictions, thus achieving robust cross-modal retrieval with query shift.
    \item Extensive experiments verify the effectiveness of the proposed method. Furthermore, we benchmark the existing TTA methods on cross-modal retrieval with query shift across six widely-used image-text datasets, hoping to facilitate the study of test-time adaptation beyond unimodal tasks.   
\end{itemize}

\section{Related Work}
\label{sec: related}
In this section, we briefly review two topics related to this
work, \textit{i.e.}, domain adaptation for cross-modal retrieval and test-time adaptation.

\subsection{Domain Adaptation for Cross-modal Retrieval}
Cross-modal retrieval aims to establish a well-structured common space, where semantically-relevant candidates could be prioritized for the queries. 
However, most existing cross-modal retrieval methods implicitly assume that the given queries follow the same distribution as the source data.
Unfortunately, such an ideal assumption is easily violated due to the complexity of real-world applications, leading to the query shift problem as discussed in Introduction. 
To address the problem, some Unsupervised Domain Adaptation (UDA) methods have been proposed to reconcile the distribution differences for robust cross-modal retrieval.
Based on the way to achieve robustness against query shift, these approaches could be roughly grouped into the following three categories: 
i) pseudo-labeling methods~\citep{UDACVR,DADA}, which first select the most relevant cross-modal pairs as positives while treating the irrelevant pairs as negatives and then conduct metric learning upon the pairs to adapt the model for target domains; 
ii) domain alignment methods~\citep{DASG}, which mitigate distribution discrepancies between the target and source domains by resorting to the maximum mean discrepancy minimization or mutual information minimization approaches; 
iii) prototype-based methods~\citep{ACP,CAPQ}, which first constructs different sets of prototypes to represent various domains and then achieves domain adaptation by minimizing the KL divergence between the corresponding prototype sets. 

Despite the promising performance, the existing domain adaptation works for cross-modal retrieval require accessing the entire target domain.
As a result, these works cannot achieve adaptation for the online query stream, limiting their practicability in real-time scenarios such as search engine.  
Different from them, this paper proposes a new adaptation method for addressing the query shift problem, which could be one of the first online adaptation approaches for cross-modal retrieval.

\subsection{Test-time adaptation}
Test-time Adaptation (TTA) has emerged as a promising avenue for domain adaptation, which aims to reconcile the distribution shifts in an online manner.
Towards achieving this goal, some Test-Time Training (TTT) approaches~\citep{TTT,TTT++,TTT1} have been proposed, which require modifying the training process of the source model and adding an auxiliary self-supervised task.
As a result, the source model could be adapted by performing the self-supervised task on the online target data stream. 
To avoid the reduplicated training cost of the source model, Fully Test-Time Adaptation~\citet{Tent} paradigm has been proposed, which could be coarsely divided into the following three categories: i) online TTA methods~\citep{TEA,COTTA}, which continually update the normalization layers by resorting to the unsupervised objectives, such as entropy minimization or its variants. ii) robust TTA methods~\citep{DeYO,TTA-Retrieval}, which strive to improve the robustness against noisy predictions, mixed distribution shifts, label shifts, and so on. iii) TTA beyond recognition, which focuses on the tasks including but not limited to image restoration~\citep{TAO}, multimodal recognition~\citep{READ}, and multimodal segmentation~\citep{MM-TTA}.

In this paper, we focus on online TTA for achieving robust cross-modal retrieval against query shift.
Among existing approaches, DISC~\citep{DISC} is most relevant to our work, while having significantly different motivations.
In brief, DISC is designed to adapt the pre-trained image hashing models against the distribution shift for achieving effective retrieval in the target domain.
Different from DISC, our work focuses on addressing the query shift problem for cross-modal retrieval, which is less-touched by the existing studies.
Moreover, unlike DISC that requires accessing the source data to train the hashing model, our TCR could adapt the off-the-shelf pre-trained models without using the source data.  

\begin{figure*}[t]
\centering
\includegraphics[width=0.98\linewidth]{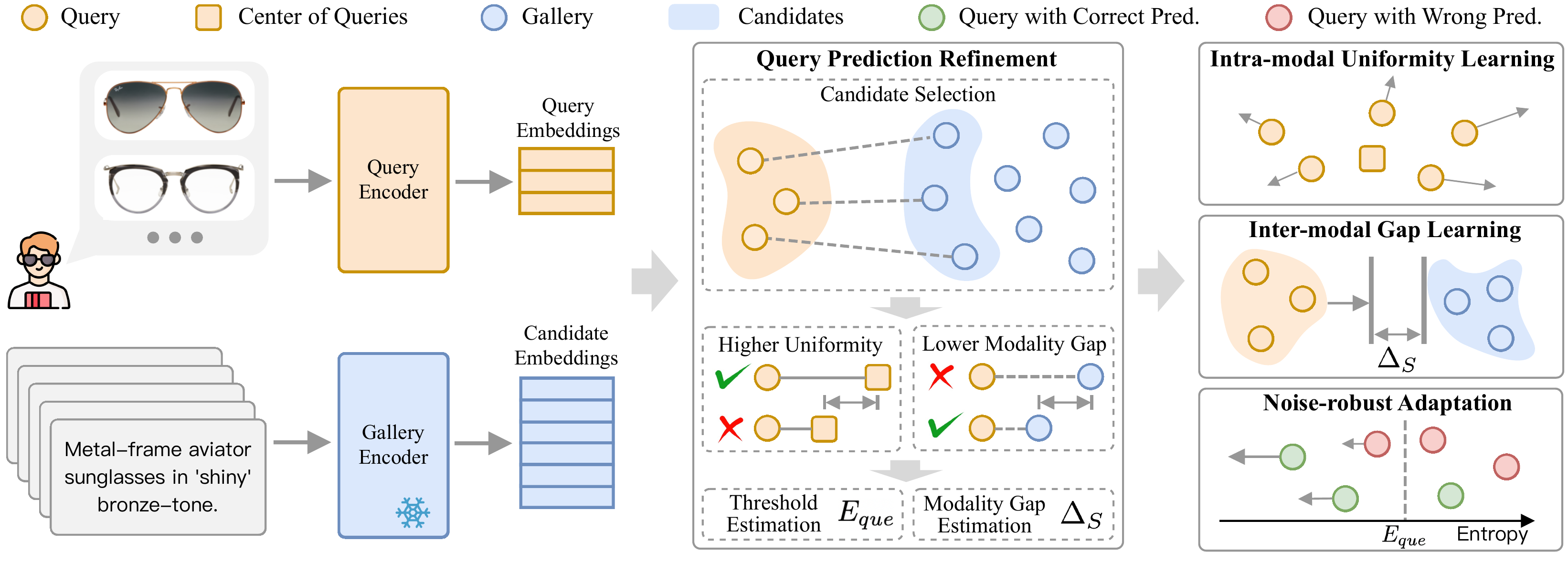}
\vspace{-0.05in}
\caption{
Overview of the proposed TCR. 
For the given online queries, the modality-specific encoders are employed to project the query and gallery samples into the latent space established by the source model. 
The obtained query embeddings and gallery embeddings are passed into the query prediction refinement module.
In the module, TCR first selects the most similar gallery sample for each query and obtain the query-gallery pairs.
After that, the pairs with higher uniformity and lower modality gap are chosen to estimate the filtering threshold of query predictions and modality gap of the source model as the constraints for the adaptation.
Finally, three loss functions are employed to achieve robust adaptation for cross-modal retrieval with query shift.
}
\vspace{-0.15in}
\label{fig: fig2}
\end{figure*}

\section{Method}
In this section, we introduce the proposed Test-time adaptation for Cross-modal Retrieval (TCR) to handle the query shift problem for cross-modal retrieval.
The section is structured as follows.
In Section~\ref{sec: problem}, we present the formal definition of the query shift problem and design a simple baseline to facilitate TTA for cross-modal retrieval.
In Section~\ref{sec: prediction refinement}, we propose the query prediction refinement module to derive the retrieval-favorable query predictions.
In Section~\ref{sec: uniformity learning}-\ref{sec: noise-robust adaptation}, we design a novel joint objective function to achieve robust cross-modal retrieval against query shift.

\subsection{Notations and Problem Formulation}
\label{sec: problem}
Without loss of generality, we take two modalities as a showcase to elaborate on the query shift problem. 
Let $f_{\Theta_{s}}$ denote the multimodal model pre-trained on the source-domain data $\mathcal{D}_{S}$, which consists of two modality-specific encoders, \textit{i.e.}, $f_{\Theta^{Q}_{s}}$ and $f_{\Theta^{G}_{s}}$.
For a given query $\mathbf{x}^{Q}_{i}$ from the target domain data  $\mathcal{D}_{T} = \left\{\mathbf{X}^{Q}=\{\mathbf{x}_{i}^{Q}\}_{i=1}^{N^{Q}}, \mathbf{X}^{G}=\{\mathbf{x}_{j}^{G}\}_{j=1}^{N^{G}}\right\}$, cross-modal retrieval aims to associate the corresponding sample $\mathbf{x}^{G}_{j}$ from the gallery set $\mathbf{X}^{G}$ by resorting to the common space established by $f_{\Theta_{s}}$, where $Q$ and $G$ denotes as query modality and gallery modality for clarity in the following.
In real-world applications, the given queries are usually from the distribution distinct with the source-domain data, \textit{i.e.}, $\mathcal{P}\left(\mathcal{D}_{T}\right) \not\sim \mathcal{P}\left(\mathcal{D}_{S}\right)$, where $\mathcal{P}\left(\mathcal{D}\right)$ denotes the distribution of the given data $\mathcal{D}$.
As a result, such a query shift problem would lead to the performance degradation of $f_{\Theta_{s}}$ as verified in the experiments.

To achieve online adaptation for $f_{\Theta_{s}}$ with query shift, we first propose a simple baseline that endows the unimodal-recognition-oriented TTA approaches with the capacity to handle the cross-modal retrieval task.
To be specific, we formulate the retrieval task as a query prediction process in analogy with the recognition task that assigns the given samples to their corresponding categories.
Formally, for a given online batch of queries with size $B$, the corresponding query predictions could be defined as
\begin{equation}
\mathbf{p}=\operatorname{Softmax}\left(\mathbf{z}^{Q}\left(\mathbf{Z}^{G}\right)^{T}/\tau \right),
    \label{eq: prediction}
\end{equation}
where $\tau$ is the temperature that controls the trade-off between the smoothness and sharpness of the predictions, $\mathbf{z}^{Q}=f_{\Theta^{Q}_{s}}(\mathbf{x}^{Q})\in \mathbb{R}^{B\times D}$ and $\mathbf{Z}^{G}=f_{\Theta^{G}_{s}}(\mathbf{X}^{G})\in \mathbb{R}^{N^{G}\times D}$ are the $\ell 2$-normalized embeddings with the dimensionality of $D$ for the given query and gallery samples, respectively. 
Thanks to the above formulation, most existing TTA methods could be adopted to handle the query shift challenge with the following objective, 
\begin{equation}
        \min_{\tilde{\Theta}} \mathcal{L}_{TTA}\left(\mathbf{p}\right),
    \label{eq: tta}
\end{equation}
where $\tilde{\Theta} \subseteq \Theta_{s}$ is the learnable parameters. 
However, such a simple formulation of query prediction would result in either model underfitting or overfitting even with carefully tuning of $\tau$ due to the variable gallery sizes, as verified in Table~\ref{tab: both-domain} and Fig.~\ref{fig: ablation}(a).
Furthermore, such a simple baseline overlooks the underlying influences behind the query shift problem and thus fails to achieve promising performance as discussed in Introduction.
On the one hand, the existing TTA methods cannot explicitly manipulate the intra-modality uniformity and inter-modality gap.
On the other hand, these methods struggle to account for the heavy noise from the query prediction.

As a remedy, we propose Test-time adaptation for Cross-modal Retrieval (TCR).
As shown in Fig.~\ref{fig: fig2}, for the given online batch of queries, TCR first employs the novel query prediction refinement module to obtain the retrieval-favorable predictions $\hat{\mathbf{p}}$.
After that, the following joint objective consisting of three novel loss functions is adopted to achieve robust cross-modal retrieval against the query shift, \textit{i.e.},
\begin{equation}
\min_{\tilde{\Theta}} \mathcal{L}\left(\hat{\mathbf{p}}\right),
    \label{eq: overall}
\end{equation}
where $\mathcal{L}= \mathcal{L}_{MU}+\mathcal{L}_{MG}+\mathcal{L}_{NA}$, with $\mathcal{L}_{MU}$, $\mathcal{L}_{MG}$, and $\mathcal{L}_{NA}$ denote the intra-modality uniformity learning loss, inter-modality gap learning loss, and the noise-robust adaptation loss, respectively. 
In the following, we will elaborate on each loss individually.
\subsection{Query Prediction Refinement}
\label{sec: prediction refinement}
In this section, we introduce the query prediction refinement module which involves refining the prediction for the given queries and estimating the constraints to support the optimization of Eq.~\ref{eq: overall}.
\subsubsection{Candidate Selection} 
To break the dilemma of vanilla query prediction formulation in Eq.~\ref{eq: prediction}, we propose selecting a subset of candidates from the gallery and then establishing new query predictions for the given queries.
Mathematically, for each given query $\mathbf{x}^{Q}_{i}$, the corresponding candidate in the gallery is obtain via 
\begin{equation}
    \mathbf{x}^{G^{\prime}}_{i}=\mathcal{N}(\mathbf{x}^{Q}_{i}),
    \label{eq: construct gallery}
\end{equation}
where $\mathcal{N}(\cdot)$ denotes the selection manner, and we adopt the nearest neighborhood selection in the common space for simplicity.
In other words, we retrieve the most similar sample from the gallery set for each query and thus obtain query-candidate pairs $(\mathbf{z}^{Q}, \mathbf{z}^{G^{\prime}})$.
Consequently, the refined query predictions for the online-batched queries $\mathbf{x}^{Q}$ could be formulated as follows,
\begin{equation}
    \mathbf{\hat{p}}=\operatorname{Softmax}\left(\mathbf{z}^{Q} \left(\mathbf{Z}^{G^{\prime}}\right)^{T}/ \tau \right),
    \label{eq: new prediction}
\end{equation}
where $\mathbf{Z}^{G^{\prime}}$ are the embeddings of the selected candidates for $\mathbf{x}^{Q}$.
The query prediction refinement manner embraces the following two merits.
On the one hand, the query prediction refinement manner could exclude some irrelevant samples in the gallery, thus preventing the model from overfitting to some extent.
On the other hand, the excluded irrelevant samples would avoid looking for a needle in a bottle of hay for queries, thus alleviating the model underfitting issue.

\subsubsection{Constraint Estimation}
It is widely acknowledged that source data could effectively regulate the domain adaptation process~\citep{JMMD,CAN}, thus circumventing the catastrophic forgetting issue.
Due to the unavailability of the source data in the test-time adaptation, we propose further choosing some source-domain-like data from the selected query-candidate pairs to estimate desirable constraints that support the optimization of Eq.~\ref{eq: overall}.
To this end, we first design a criterion for choosing the query-candidate pair $(\mathbf{z}_{i}^{Q}, \mathbf{z}_{i}^{G^{\prime}})$, \textit{i.e.},
\begin{equation}
    \text{SI}=2\left(\|\mathbf{z}_{i}^{Q}-\mathbf{z}_{i}^{G^{\prime}}\|\right)-\left(\|\mathbf{z}_{i}^{Q}-\overline{\mathbf{z}}^Q\|+\|\mathbf{z}_{i}^{G^{\prime}}-\overline{\mathbf{z}}^{G^{\prime}}\|\right),
    \label{eq: SI}
\end{equation}
where $\overline{\mathbf{z}}^Q=\frac{1}{B}\sum_{i}^{B}\mathbf{z}_i^Q$ and $\overline{\mathbf{z}}^{G^{\prime}}=\frac{1}{B}\sum_{i}^{B}\mathbf{z}_i^{G^{\prime}}$ are the centers of the given queries and the selected candidates, respectively.
In the implementation, we choose $30\%$ query-candidate pairs with the smallest $\text{SI}$ value, namely, $(\mathbf{z}^{Q_m}, \mathbf{z}^{G^{\prime}_m})$ of size $M$, as the source-domain-like data.
Clearly, a low value of the criterion has the incentive to select the query-candidate pairs with the small modality gap and high intra-modality uniformity, which have higher probability to be source-domain-like, as verified in Fig.~\ref{fig: analytic-two-charac}.  

Based on the selected source-domain-like data, we propose estimating the modality gap of the source model as follows,
\begin{equation}
    \Delta_{S}=\left\| \frac{1}{M}\sum_{i}^{M}\mathbf{z}^{Q_{m}}_{i}- \frac{1}{M}\sum_{j}^{M}\mathbf{z}^{G_{m}^{\prime}}_{j} \right\|,
    \label{eq: estimate mg}
\end{equation}
where $\mathbf{z}^{Q_{m}}_{i}$ and $\mathbf{z}^{G_{m}^{\prime}}_{j}$ denote the $i$-th query sample and the $j$-th gallery sample in the query-candidate pairs, respectively.
As the another by-product, a desirable threshold that could filter the noise in the query predictions $\hat{\mathbf{p}}$ could be adaptively determined as follows,
\begin{equation}
    E_{m}=\max_{i=1, \dots, M}  E\left( \mathbf{x}_{i}^{Q_{m}} \right) ,
\label{eq: get_max_entropy}
\end{equation}
where $E\left(\cdot\right)$ indicates the entropy based on the refined query predictions.

\subsection{Intra-modality Uniformity Learning}
\label{sec: uniformity learning}
As discussed in Introduction, query shift would diminish the uniformity of the query modality, resulting in confused queries with lower discrimination in the common space.
To address the problem, we propose to perform the contrast between queries and their respective centers, thus explicitly enlarging the intra-modality uniformity.
Mathematically, the loss function of intra-modality uniformity learning is defined as follows,
\begin{equation}
    \mathcal{L}_{MU} = \frac{1}{B}\sum_{i}^{B}exp\left(-\|\mathbf{z}_i^Q - \overline{\mathbf{z}}^Q\|/t \right)
    \label{eq: loss uniformity}
\end{equation}
where $t$ is the trade-off parameter to control the uniformity that is fixed as $10$ in the experiments.

\subsection{Inter-modality Gap Leaning}
\label{sec: modality gap learning}
As discussed in Introduction, query shift would amplify the modality gap between query and gallery modalities, disrupting the cross-modal alignment established by the source model.
To remedy this, we propose to rectify the difference between the query and gallery modalities to the estimated modality gap of the source model in Eq.~\ref{eq: estimate mg}.
Formally, the inter-modality gap learning loss is defined as follows,
\begin{equation}
    \mathcal{L}_{MG}=\left(\Delta_{T} - \Delta_{S} \right)^{2},
    \label{eq: loss_mmg}
\end{equation}
where $\Delta_{T}=\left\|\overline{\mathbf{z}}^Q-\overline{\mathbf{z}}^{G^{\prime}}\right\|$ denotes the modality gap of the target domain. 
The key idea behind $\mathcal{L}_{MG}$ is that the modality gap rectification would take advantage of well-aligned multimodal common space from the source model, thus boosting retrieval performance.
Notably, as observed in~\citet{MindGap}, over-eliminating the modality gap would not improve or even degrade the performance of the multimodal model.
Therefore, we believe the proposed loss that rectifies the modality gap of the target model to a plausible constraint in a non-monotonic manner is reasonable.
The experimental results in Fig.~\ref{fig: analytic-two-charac} could support the claims.

\subsection{Noise-robust Adaptation}
\label{sec: noise-robust adaptation}
To achieve robustness against the heavy noise on the query predictions, we propose the following noise-robust adaptation loss,
\begin{equation}
    \mathcal{L}_{NA}=\frac{1}{\sum_{i} \mathbb{I}_{\{S(\mathbf{x}_{i}^{Q})\neq 0\}}} \sum_{i=1}^{N^{Q}} S(\mathbf{x}_{i}^{Q}) E(\mathbf{x}_{i}^{Q}), \ \text{where} \ S(\mathbf{x}_{i}^{Q})=\max \left(1-\frac{E(\mathbf{x}_{i}^{Q})}{E_{m}}, 0\right),
    \label{eq: loss_rem}
\end{equation}
where $E_{m}$ is the self-adaptive threshold estimated in Eq.~\ref{eq: get_max_entropy}, and $\mathbb{I}_{\{\cdot\}}$ is an indicator function evaluating to $1$ \textit{i.f.f.} the condition is satisfied.
It is worth noting that existing TTA methods like EATA~\citep{EATA} and SAR~\citep{SAR} are highly sensitive to the manually determined thresholds, resulting in either none or all query predictions being treated as noise.
In contrast, the proposed $\mathcal{L}_{NA}$ not only employs a self-adaptive threshold to filter out noise but also employs various confidence scores for query predictions to facilitate the optimization.

\section{Experiments}
In this section, we verify the effectiveness of TCR in handling the query shift problem for the image-text retrieval task.
This section is organized as follows.
In Section~\ref{sec: experiment setting}, we present the implementation details and experiment settings of TCR.
In Section~\ref{sec: comparisons SOTA}, we conduct extensive comparison experiments to verify the performance superiority of TCR.
In Section~\ref{sec: ablation}, we perform a series of analytic studies, ablation studies, and visualization analyses, to provide a comprehensive understanding of TCR.
\begin{table*}[t]
    \vspace{-0.3in}
    \caption{Comparisons with state-of-the-art methods on COCO-C benchmark under \textbf{\textsc{query shift on the image modality}} with maximum severity level regarding the Recall@1 metric. The best results are marked in \textbf{bold}.   
    }
    \vspace{-0.05in}
    \label{tab: coco-c-image}
\newcommand{\tabincell}[2]{\begin{tabular}{@{}#1@{}}#2\end{tabular}}
 \begin{center}
 \begin{threeparttable}
 \LARGE
    \resizebox{0.98\linewidth}{!}{
 	\begin{tabular}{l|cccc|cccc|cccc|cccc|>{\columncolor{blue!8}}c}
 	\multicolumn{1}{c}{} & \multicolumn{4}{c}{Noise} & \multicolumn{4}{c}{Blur} & \multicolumn{4}{c}{Weather} & \multicolumn{4}{c}{Digital}  \\
 	 Query Shift & Gauss. & Shot & Impul. &Speckle & Defoc. & Glass & Motion & Zoom & Snow & Frost & Fog & Brit. & Contr. & Elastic & Pixel & JPEG & Avg.  \\
    \cmidrule{1-18}
        BLIP ViT-B/16 &  43.4 & 46.3 & 43.2 & 57.3 & 43.3 & 68.0 & 39.7 & 8.4 & 32.3 & 52.2 & 57.0 & 66.8 & 36.0 & 41.3 & 20.6 & 63.7 & 45.0 \\ 
        ~~$\bullet~$Tent & 41.6 & 40.5 & 37.9 & 54.0 & 44.7 & 65.1 & 39.6 & 8.3  & 31.9 & 48.7 & 56.3 & 66.5 & 31.8 & 40.3 & 19.2 & 62.3 & 43.0 \\ %
        ~~$\bullet~$EATA & 41.4 & 50.3 & 35.7 & 63.1 & 49.8 & 72.2 & 46.2 & 6.9  & 45.6 & 56.7 & 62.5 & 71.4 & 43.6 & 51.3 & 25.6 & 67.0 & 49.3  \\
        ~~$\bullet~$SAR & 42.3 & 51.5 & 37.5 & 61.8 & 40.3 & 71.5 & 32.8 & 6.2  & 38.0 & 56.2 & 59.1 & 70.6 & 31.1 & 53.5 & 17.5 & 66.4 & 46.0  \\
        ~~$\bullet~$READ & 45.8 & 48.4 & 37.2 & 59.9 & 44.5 & 71.8 & 46.6 & 11.5 & 39.9 & 49.9 & 58.4 & 70.3 & 35.8 & 45.0 & 18.8 & 66.2 & 46.9  \\
        ~~$\bullet~$DeYO & 47.9 & 53.5 & 46.8 & 63.4 & 42.9 & 72.1 & 36.7 & 3.2  & 37.5 & 59.7 & 66.4 & 71.2 & 40.3 & 49.0 & 13.1 & 67.6 & 48.2 \\
        \rowcolor{pink!30}~~$\bullet~$Ours & \textbf{53.2} & \textbf{56.2} & \textbf{54.8} & \textbf{64.6} & \textbf{58.0} & \textbf{73.7} & \textbf{56.4} & \textbf{32.2} & \textbf{56.5} & \textbf{64.1} & \textbf{71.0} & \textbf{73.4} & \textbf{57.9} & \textbf{63.7} & \textbf{41.8} & \textbf{68.4} & \textbf{59.1} \\
    \cmidrule{1-18}
        BLIP ViT-L/16 & 50.3 & 51.8 & 51.1 & 61.6 & 53.7 & 72.1 & 49.4 & 14.5 & 44.0 & 57.5 & 61.8 & 70.5 & 37.3 & 50.6 & 32.0 & 70.5 & 51.8 \\ 
        ~~$\bullet~$Tent & 46.3 & 49.3 & 46.7 & 58.4 & 52.2 & 71.8 & 47.5 & 12.3 & 41.9 & 56.2 & 60.9 & 69.7 & 35.7 & 48.3 & 29.4 & 69.6 & 49.8  \\ %
        ~~$\bullet~$EATA & 46.2 & 53.5 & 49.5 & 63.8 & 56.5 & 73.8 & 52.6 & 18.4 & 50.6 & 59.1 & 64.5 & 72.1 & 40.7 & 55.4 & 43.5 & 70.7 & 54.4 \\
        ~~$\bullet~$SAR & 45.9 & 50.2 & 47.3 & 63.1 & 51.1 & 73.8 & 47.2 & 11.6 & 40.8 & 58.9 & 60.7 & 71.6 & 33.6 & 54.0 & 34.4 & 70.5 & 50.9  \\
        ~~$\bullet~$READ & 38.1 & 48.0 & 43.3 & 63.5 & 43.6 & 73.4 & 43.6 & 22.0 & 44.5 & 56.5 & 62.2 & 71.9 & 32.9 & 49.6 & 27.5 & 70.6 & 49.5 \\
        ~~$\bullet~$DeYO & 39.9 & 50.2 & 43.5 & 63.8 & 50.4 & 74.0 & 52.4 & 5.4 & 49.5 & 59.3 & 62.8 & 71.8 & 34.0 & 54.7 & 34.4 & 69.7 & 51.0  \\
        \rowcolor{pink!30}~~$\bullet~$Ours & \textbf{58.2} & \textbf{60.7} & \textbf{59.8} & \textbf{66.6} & \textbf{61.5} & \textbf{74.9} & \textbf{60.3} & \textbf{36.8} & \textbf{59.0} & \textbf{65.2} & \textbf{72.1} & \textbf{73.5} & \textbf{56.3} & \textbf{65.7} & \textbf{50.2} & \textbf{71.6} & \textbf{62.0} \\
    \cmidrule{1-18}
	\end{tabular}
	}
	 \end{threeparttable}
	 \end{center}
\vspace{-0.2in}
\end{table*}

\begin{table*}[t]
    \vspace{-0.1in}
    \caption{Comparisons with state-of-the-art methods on COCO-C benchmark under \textbf{\textsc{query shift on the text modality}} with maximum severity level regarding the Recall@1 metric.
    }
    \vspace{-0.05in}
    \label{tab: coco-c-text}
\newcommand{\tabincell}[2]{\begin{tabular}{@{}#1@{}}#2\end{tabular}}
 \begin{center}
 \begin{threeparttable}
 \LARGE
    \resizebox{0.90\linewidth}{!}{
 	\begin{tabular}{l|ccccc|ccccc|ccccc|>{\columncolor{blue!8}}c}
 	\multicolumn{1}{c}{} & \multicolumn{5}{c}{Character-level} & \multicolumn{5}{c}{Word-level} & \multicolumn{5}{c}{Sentence-level}  \\
 	 Query Shift & OCR & CI & CR & CS & CD & SR & RI & RS & RD & IP & Formal & Casual & Passive & Active & Backtrans & Avg.  \\
    \cmidrule{1-17}
        BLIP ViT-B/16 & 31.4 & 11.3 & 9.4 & 18.9 & 11.4 & 43.6 & 51.5 & 50.3 & 50.6 & 56.8 & 56.6 & 56.2 & 54.9 & 56.8 & 54.2 & 40.9 \\ 
        ~~$\bullet~$Tent & 31.4 & 11.0 & 9.5  & 17.7 & 11.3 & 43.2 & 51.3 & 50.3 & 50.6 & 56.6 & 56.2 & 56.0 & 54.9 & 56.9 & 53.9 & 40.7    \\ %
        ~~$\bullet~$EATA & 33.1 & 11.9 & 10.5 & 18.4 & 12.0 & 44.9 & 53.0 & 51.6 & 50.3 & 56.2 & 56.8 & \textbf{56.8} & \textbf{56.0} & 56.8 & 54.3 & 41.5   \\
        ~~$\bullet~$SAR & 31.8 & 11.6 & 9.9  & 18.5 & 11.7 & 43.6 & 51.5 & 50.3 & 50.6 & 56.8 & 56.5 & 56.2 & 54.9 & 56.8 & 54.2 & 41.0     \\
        ~~$\bullet~$READ & 32.3 & 11.4 & 9.6  & 18.2 & 11.2 & 44.3 & 52.9 & 51.7 & 51.1 & 57.6 & 57.1 & 56.7 & 55.9 & 57.1 & \textbf{54.7} & 41.4    \\
        ~~$\bullet~$DeYO & 31.4 & 11.3 & 9.4  & 17.9 & 11.4 & 43.6 & 51.5 & 50.3 & 50.6 & 56.8 & 56.5 & 56.2 & 54.9 & 56.7 & 54.2 & 40.9    \\
        \rowcolor{pink!30}~~$\bullet~$Ours & \textbf{34.1} & \textbf{13.7} & \textbf{11.8} & \textbf{19.5} & \textbf{13.2} & \textbf{45.3} & \textbf{53.8} & \textbf{51.8} & \textbf{51.5} & \textbf{57.3} & \textbf{57.1} & \textbf{56.8} & \textbf{56.0} & \textbf{57.3} & \textbf{54.7} & \textbf{42.3} \\
    \cmidrule{1-17}
        BLIP ViT-L/16 & 34.5 & 12.3 & 11.1 & 19.7 & 12.9 & 46.0 & 54.4 & 54.0 & 53.5 & 59.4 & 59.1 & 58.8 & 57.8 & 59.4 & 56.7 & 43.3 \\ 
        ~~$\bullet~$Tent & 34.0 & 12.3 & 11.0 & 19.6 & 12.9 & 46.5 & 54.2 & 53.8 & 53.4 & 59.4 & 59.1 & 58.8 & 57.6 & 58.9 & 56.5 & 43.2    \\ %
        ~~$\bullet~$EATA & 35.6 & 13.3 & 11.3 & 20.3 & 13.2 & 47.2 & 55.4 & 54.2 & 53.8 & 59.2 & 59.1 & 59.4 & 57.9 & 59.4 & 56.8 & 43.7  \\
        ~~$\bullet~$SAR & 34.5 & 13.1 & 11.2 & 20.3 & 13.1 & 46.7 & 54.4 & 54.0 & 53.5 & 59.5 & 59.1 & 58.8 & 57.8 & 59.4 & 56.7 & 43.5    \\
        ~~$\bullet~$READ & 35.3 & 12.2 & 10.9 & 19.1 & 12.7 & 47.3 & 55.1 & 55.0 & 53.3 & 59.7 & 59.3 & \textbf{59.1} & 58.1 & \textbf{59.6} & 56.7 & 43.6    \\
        ~~$\bullet~$DeYO & 34.5 & 12.3 & 11.1 & 19.7 & 12.9 & 46.7 & 54.4 & 54.0 & 53.5 & 59.5 & 59.1 & 58.8 & 57.8 & 59.4 & 56.7 & 43.4   \\
        \rowcolor{pink!30}~~$\bullet~$Ours & \textbf{36.8} & \textbf{14.7} & \textbf{13.4} & \textbf{21.3} & \textbf{14.3} & \textbf{47.9} & \textbf{56.3} & \textbf{54.8} & \textbf{53.9} & \textbf{59.5} & \textbf{59.4} & 59.0 & \textbf{58.2} & \textbf{59.6} & \textbf{56.9} & \textbf{44.4} \\
    \cmidrule{1-17}
	\end{tabular}
	}
	 \end{threeparttable}
	 \end{center}
\vspace{-0.2in}
\end{table*}

\subsection{Implementation Details and Experiment Settings}
\label{sec: experiment setting}
TCR is a general TTA framework that could endow most existing pre-trained models with robustness against the query shift.
Therefore, we select CLIP~\citep{CLIP} and BLIP~\citep{BLIP} as the source models since they are the widely adopted vision-language models for image-text retrieval task.
Following~\citet{SCAN}, we adopt two testing protocols, namely, image-to-text retrieval (a.k.a. TR) and text-to-image retrieval (a.k.a. IR).
During the adaptation process, TCR performs the objective function for each coming mini-batch of queries, and the batch size is set as $64$. 
Following~\citet{SAR, Tent}, TCR updates the parameters within the normalization layers in the query-specific encoder $f_{\Theta^{Q}_{s}}$using the AdamW optimizer. 
To be more specific, the learnable parameters in $\tilde{\Theta}$ (Eq.~\ref{eq: overall}) correspond to the Layer Normalization (LN) layers in our implementation.
Besides, the temperature hyper-parameter $\tau$ in Eq.~\ref{eq: prediction} and uniformity learning hyper-parameter $t$ in Eq.~\ref{eq: loss uniformity} are fixed as $0.02$ and $10$ for all experiments, respectively.

To investigate the influence of cross-modal retrieval with query shift, we employ the following two settings for extensive evaluations (see more details in Appendix~\ref{Appendix: benchmark details}). 
\begin{itemize}
    \item \textbf{Query Shift} (QS):  
    In this setting, only the queries come from different distributions with the source-domain data. 
    Following \citet{MMbenchmark}, we introduce 16 types of corruptions to the image modality and 15 types to the text modality across widely-used image-text retrieval datasets, COCO~\citep{COCO} and Flickr~\citep{Flickr}. 
    As a result, the COCO-C and Flickr-C benchmarks are constructed, which would result in distribution shifts on either the image or text modalities.
    To guarantee the controlled study on QS, we first fine-tune the pre-trained model on either the COCO~\citep{COCO} or Flickr~\citep{Flickr} dataset, namely, treating them as the source domains.
    After that, evaluations are conducted on the COCO-C or Flickr-C benchmarks, namely, treating them as the target domain.
 \item \textbf{Query-Gallery Shift} (QGS): 
    In this setting, both the query and gallery samples are drawn from distributions different from the source-domain data.
    To this end, evaluations are directly conducted on the pre-trained model upon several widely-used image-text retrieval datasets from various domains, including Fashion-Gen~\citep{Fashion-Gen} from the e-commerce domain, CUHK-PEDES~\citep{CUHK} and ICFG-PEDES~\citep{ICFG} from the person re-identification (ReID) domain, and COCO, Flickr, and Nocaps~\citep{Nocaps} from the natural image domain.
    In other words, the source model would encounter distribution shifts on both image and text modalities during adaptation.
\end{itemize}

\begin{table*}[t]
    \vspace{-0.15in}
    \caption{Comparisons with state-of-the-art methods on benchmarks under \textbf{\textsc{Query-Gallery shifts}} regarding the Recall@1 metric. In the table, ``ID", ``ND" and ``OD" refer to ``In-Domain", ``Near-Domain" and ``Out-Domain", respectively.
    Besides, ``TR@1" / ``IR@1" represent Recall@1 for image-to-text retrieval / text-to-image retrieval.
    }
    \vspace{-0.05in}
    \label{tab: both-domain}
\newcommand{\tabincell}[2]{\begin{tabular}{@{}#1@{}}#2\end{tabular}}
 \begin{center}
\begin{threeparttable}  
\LARGE     
\resizebox{0.9\linewidth}{!}{  
\begin{tabular}{l|cc|cc|cc|cc|cc|cc|>{\columncolor{blue!8}}c}  
\multicolumn{1}{c}{} & \multicolumn{2}{c}{Base2Flickr} & \multicolumn{2}{c}{Base2COCO} & \multicolumn{2}{c}{Base2Fashion} & \multicolumn{2}{c}{Base2Nocaps(ID)} & \multicolumn{2}{c}{Base2Nocaps(ND)} & \multicolumn{2}{c}{Base2Nocaps(OD)} \\  
Query Shift & $\text{TR@1}$ & $\text{IR@1}$ & $\text{TR@1}$ & $\text{IR@1}$ & $\text{TR@1}$ & $\text{IR@1}$ & $\text{TR@1}$ & $\text{IR@1}$ & $\text{TR@1}$ & $\text{IR@1}$ & $\text{TR@1}$ & $\text{IR@1}$ & Avg.  \\    
\cmidrule{1-14}         
CLIP ViT-B/16 & 80.2 & 61.5 & 52.5 & 33.0 & 8.5  & 13.2 & 84.9 & 61.4 & 75.4 & 49.2 & 73.8 & 55.8 & 54.1 \\
~~$\bullet~$Tent & 81.4                         & 64.0 & 48.8 & 27.6 & 5.6  & 10.7 & 85.1 & 61.7 & 74.6 & 48.6 & 71.8 & 56.1 & 53.0 \\
~~$\bullet~$EATA & 80.4                         & 63.4 & 52.1 & 34.8 & 8.1  & 12.0 & 84.7 & 62.0 & 75.1 & 52.3 & 74.1 & 56.9 & 54.7 \\
~~$\bullet~$SAR & 80.3                         & 62.2 & 51.8 & 33.9 & 8.0  & 13.3 & 84.7 & 61.3 & 75.4 & 51.3 & 73.7 & 56.1 & 54.3 \\
~~$\bullet~$READ & 80.6                         & 64.4 & 46.0 & 35.7 & 5.8  & 11.2 & 85.1 & 63.0 & 75.0 & 52.1 & 73.5 & 57.0 & 54.1 \\
~~$\bullet~$DeYO & 80.1                         & 64.0 & 51.5 & 33.4 & 6.9  & 10.9 & 84.4 & 62.2 & 75.1 & 52.0 & 73.2 & 57.3 & 54.3 \\
\rowcolor{pink!30}~~$\bullet~$Ours & \textbf{82.4}                         & \textbf{64.8} & \textbf{52.9} & \textbf{36.5} & \textbf{8.9}  & \textbf{14.0} & \textbf{85.1} & \textbf{63.5} & \textbf{75.7} & \textbf{54.0} & \textbf{74.4} & \textbf{58.0} & \textbf{55.9} \\
\cmidrule{1-14} 
BLIP ViT-B/16 & 70.0                         & 68.3 & 59.3 & 45.4 & 19.9 & 26.1 & 88.2 & 74.9 & 79.3 & 63.6 & 81.9 & 67.8 & 62.1 \\
~~$\bullet~$Tent & 81.9                         & 68.5 & 61.7 & 41.7 & 14.1 & 26.1 & 88.5 & 75.4 & 82.6 & 64.1 & 82.7 & 68.9 & 63.0 \\
~~$\bullet~$EATA & 82.3                         & 69.4 & 64.2 & 47.9 & 12.8 & 25.2 & 87.8 & 75.1 & 82.8 & 63.9 & 81.5 & 67.9 & 63.4 \\
~~$\bullet~$SAR & 81.7                         & 68.3 & 63.5 & 46.6 & 17.9 & 26.1 & 88.2 & 75.6 & 81.0 & 65.4 & 81.2 & 69.3 & 63.7 \\
~~$\bullet~$READ & 80.0                         & 69.9 & 62.1 & 46.4 & 5.6  & 24.1 & 87.3 & 75.1 & 80.6 & 63.9 & 80.7 & 67.9 & 62.0 \\
~~$\bullet~$DeYO & 83.5                         & 69.9 & 65.0 & 47.3 & 12.2 & 24.1 & 89.2 & 75.6 & 83.7 & 65.7 & 84.3 & 69.4 & 64.2 \\
\rowcolor{pink!30}~~$\bullet~$Ours & \textbf{86.8}                         & \textbf{70.3} & \textbf{68.9} & \textbf{48.9} & \textbf{23.6} & \textbf{30.3} & \textbf{89.7} & \textbf{76.0} & \textbf{86.3} & \textbf{66.1} & \textbf{87.2} & \textbf{69.5} & \textbf{67.0} \\
\cmidrule{1-14} 
\end{tabular}  
}  
\end{threeparttable}
	 \end{center}
\vspace{-0.25in}
\end{table*}

\subsection{Comparisons with State-of-the-arts}
\label{sec: comparisons SOTA}
In this section, We compare TCR with five SOTA TTA methods (Tent~\citep{Tent}, EATA~\citep{EATA}, SAR~\citep{SAR}, READ~\citep{READ}, and DeYO~\citep{DeYO}) under both the QS and QGS settings.
Among the baseline methods, Tent is the vanilla TTA approach with an entropy-based objective, while the others enhance Tent by incorporating specially designed noise-robust loss functions.
For a fair comparison, we select the optimal temperature (\red{Eq.1}) for the TTA baselines upon each dataset according to Fig.~\ref{fig: ablation}(a).
The results on the QS setting and QGS setting are summarized in Tables~\ref{tab: coco-c-image}, \ref{tab: coco-c-text}, \ref{tab:flickr-c-image-domain-shift}, \ref{tab:flickr-c-text-domain-shift} and Tables~\ref{tab: both-domain}, \ref{tab: reid}, respectively.
From the results, one could have the following observations and conclusions.
\begin{itemize}
    \item Existing TTA methods only achieve marginal performance improvements over the base model, which could be attributed to the inability on manipulating both modality uniformity and the modality gap.
    In contrast, TCR could rectify the modality gap and enlarge the modality uniformity, thus significantly outperforming all the baselines across various pre-trained model types and sizes.
    \item TCR demonstrates greater robustness against more severe shift types like ``Zoom" and ``Pixel" (see Table~\ref{tab: coco-c-image} and Table~\ref{tab:flickr-c-image-domain-shift}), whereas most baseline methods experience significant performance degradation under these challenging distribution shifts. 
    Moreover, in Table~\ref{tab: both-domain}, the more significant performance improvements on ``Base2Nocaps" with ``ND" and ``OD" compared to that with ``ID" also verify the conclusion.
    \item As the size of the gallery set increases in Table~\ref{tab: both-domain} (from “Base2Flickr” to “Base2COCO” to “Base2Fashion”), existing TTA methods suffer from increasing performance degradation, eventually performing even worse than the base model. This supports our claim in Section~\ref{sec: problem} that TTA methods struggle to accommodate well for cross-modal retrieval tasks due to the excessively large gallery set.
    In contrast, our method consistently achieves performance improvements across various gallery sizes.
\end{itemize}

\begin{wraptable}{r}{0.45\textwidth}
\vspace{-0.6in}
    \caption{
    Comparisons with state-of-the-art methods on ReID benchmarks under \textbf{\textsc{Query-Gallery shifts}} regarding the Recall@1 metric.
    }
    \label{tab: reid}
\newcommand{\tabincell}[2]{\begin{tabular}{@{}#1@{}}#2\end{tabular}}
 \begin{center}
\begin{threeparttable}
\LARGE
\resizebox{0.9\linewidth}{!}{
\begin{tabular}{l|c|c|>{\columncolor{blue!8}}c}
\multicolumn{1}{c}{} & \multicolumn{1}{c}{CUHK2ICFG} & \multicolumn{1}{c}{ICFG2CUHK} & \multicolumn{1}{c}{} \\  
 Query Shift & $\text{IR@1}$ & $\text{IR@1}$ & $\text{Avg.}$ \\    
\cmidrule{1-4}         
CLIP ViT-B/16 & 33.3 & 41.0 & 37.2 \\       
~~$\bullet~$Tent & 33.5 & 41.9 & 37.7  \\ %
~~$\bullet~$EATA & 33.3 & 42.2 & 37.8 \\         
~~$\bullet~$SAR & 33.3 & 42.2 & 37.8 \\   
~~$\bullet~$READ & 33.0 & 42.3 &  37.7 \\
~~$\bullet~$DeYO & 33.3 & 42.2 & 37.8 \\   
\rowcolor{pink!30}~~$\bullet~$Ours & \textbf{37.3} & \textbf{42.4} & \textbf{39.9} \\  
\cmidrule{1-4} 
\end{tabular}  
}  
\end{threeparttable}
	 \end{center}
\vspace{-0.25in}
\end{wraptable}

\subsection{Ablation and Analytic Study}
\label{sec: ablation}
In this section, all the experiments are conducted under the ``Base2COCO'' setting using the BLIP ViT-B/16 model unless otherwise stated.

\textbf{Analytic Studies on Intra-modality Uniformity and Inter-modality Gap.}
As pointed out in Introduction, the query shift would diminish the intra-modality uniformity and amplify the inter-modality gap.
For an in-depth understanding, we conduct analytic experiments to investigate how the two characteristics affect the retrieval performance.
To examine the influence of the intra-modality uniformity, we manually scale the latent distance between different queries.
Mathematically, the scaling operation is defined as follows.
\begin{equation}
    (\mathbf{z}_i^Q)^{\text {scale }}=\overline{\mathbf{Z}}^{Q}+\lambda^{\text{scale}}\left(\mathbf{z}_i^Q-\overline{\mathbf{Z}}^{Q}\right),
    \label{eq: scale}
\end{equation}

\begin{wrapfigure}{r}{0.5\textwidth}
\vspace{-0.2in}
\centering
\includegraphics[width=1.0\linewidth]{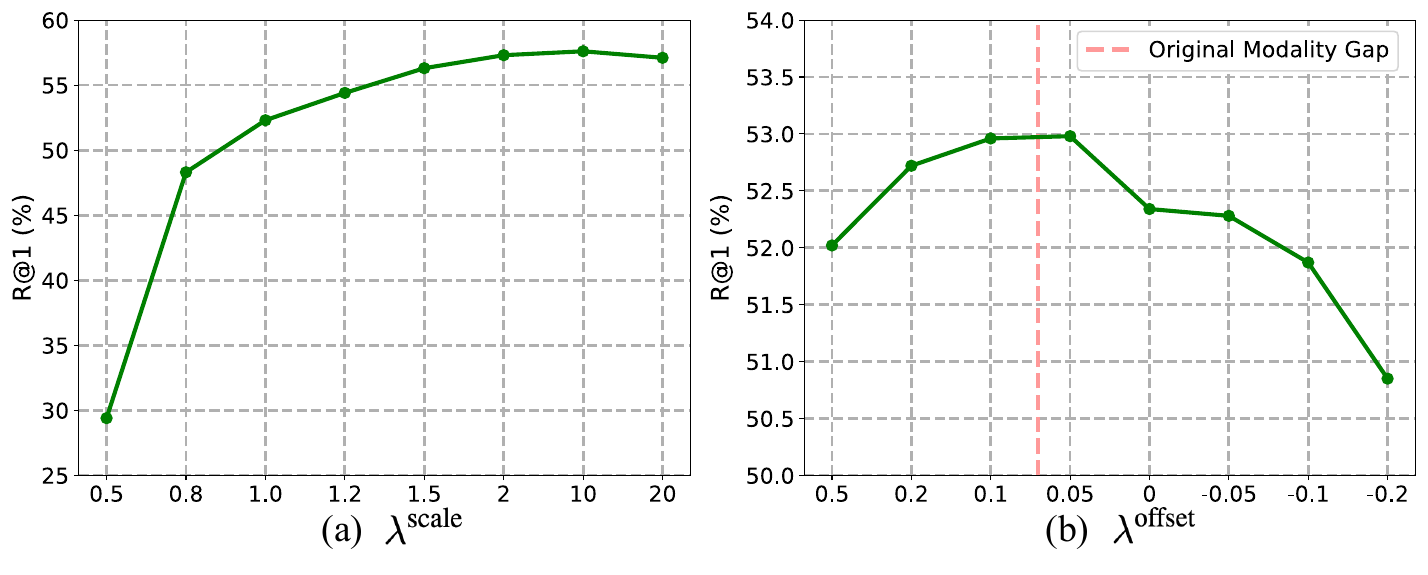}
\vspace{-0.25in}
\caption{Observation of the intra-modality uniformity and inter-modality gap. 
The increasing $\lambda^{\operatorname{scale}}$ indicates the growing intra-modality uniformity while the decreasing $\lambda^{\operatorname{offset}}$ indicates the narrowing inter-modality gap.
Notably, $\lambda^{\operatorname{scale}}=1.0$ and $\lambda^{\operatorname{offset}}=0$ represent no scaling and no offset, respectively. 
}
\vspace{-0.1in}
\label{fig: analytic-two-charac}
\end{wrapfigure}

\vspace{-0.14in}

where $\overline{\mathbf{Z}}^{Q}=\frac{1}{N^{Q}}\sum_{i=1}^{N^{Q}}\mathbf{z}_i^Q$ is the center of queries, $\lambda^{\text{scale}}$ is the scaling factor.
As illustrated in Fig.~\ref{fig: analytic-two-charac}(a), 
increasing the intra-modality uniformity in the query modality would improve the performance, but not vice versa.
Such a phenomenon indicates that higher uniformity would guarantee the discrimination between queries and thus boost performance.
To examine the influence of the inter-modality gap, following~\citet{MindGap}, we manually move every query embedding towards closing the modality gap. Formally,
\begin{equation}
    (\mathbf{z}_i^Q)^{\text {offset}}=\mathbf{z}_i^Q-\lambda^{\text {offset}} \left(\overline{\mathbf{Z}}^{Q}-\overline{\mathbf{Z}}^{G}\right),
    \label{eq: offset}
\end{equation}
where $\overline{\mathbf{Z}}^{G}=\frac{1}{N^{G}}\sum_{i=1}^{N^{G}}\mathbf{z}_i^G$ is the center of the gallery samples, $\lambda^{\text{offset}}$ controls the offset of the embeddings. 
From the results in Fig.~\ref{fig: analytic-two-charac}(b), one could observe that monotonously eliminating the modality gap would not always improve the performance. 
In contrast, the estimated modality gap from the pre-trained model is a plausible criterion for modality gap rectification.
Note that the embeddings are all $\ell 2$-normalized after scaling or shifting.

\begin{wraptable}{r}{0.43\textwidth}
\vspace{-0.3in}
    \caption{Ablation study of the loss functions, where " $\checkmark$ " denotes the loss is adopted.
    }
    \label{tab: loss ablation}
\newcommand{\tabincell}[2]{\begin{tabular}{@{}#1@{}}#2\end{tabular}}
 \begin{center}
 \begin{threeparttable}
    \resizebox{1.0\linewidth}{!}{
 	\begin{tabular}{cccccc}
            $\mathcal{L}_{NA}$   & $\mathcal{L}_{MU}$ & $\mathcal{L}_{EMG}$ & $\text{TR@1}$ & $\text{IR@1}$ & Avg. \\
            \cmidrule{1-6}
             &  &   & 59.3 & 45.4 & 52.4\\
            $\checkmark$ &  &  & 67.4 & 47.8 & 57.9\\
             & $\checkmark$ &  & 64.9 & 46.7 & 55.8\\
             &  & $\checkmark$ & 64.3 & 46.3 & 55.3\\
            $\checkmark$ & $\checkmark$ &  & 67.8 & 48.3 & 58.2\\
            $\checkmark$ &  & $\checkmark$ & 68.1 & 48.4 & 58.4\\
             & $\checkmark$ & $\checkmark$ & 66.3 & 47.8 & 57.1\\
            $\checkmark$ & $\checkmark$ & $\checkmark$ & 68.9 & 48.9 & 58.9\\
 	\midrule
	\end{tabular}
	}
	 \end{threeparttable}
	 \end{center}
\vspace{-0.1in}
\end{wraptable}

\begin{figure*}[t]
\centering
\vspace{-0.1in}
\includegraphics[width=0.9\linewidth]{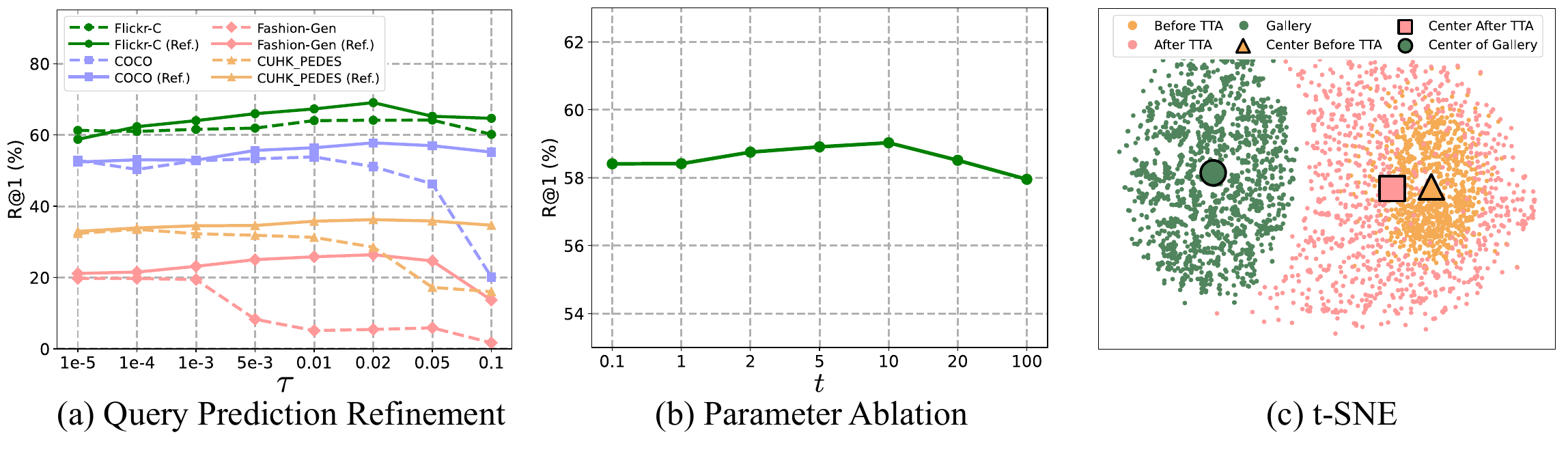}
\vspace{-0.1in}
\caption{Finer-grained Ablation studies. (a) The parameter analysis of $\tau$ (Eq.~\ref{eq: prediction} and Eq.~\ref{eq: new prediction}) on the vanilla TTA method Tent w/ (solid line) and w/o (dotted line) the query prediction refinement module.  (b) The parameter analysis of $t$ in Eq.~\ref{eq: loss uniformity}. (c) The t-SNE visualization of TR on the query and gallery embeddings after employing the proposed TCR.}
\vspace{-0.1in}
\label{fig: ablation}
\end{figure*}

\textbf{Ablation studies.} 
To verify the effectiveness of each design, we investigate the loss terms of TCR in Table~\ref{tab: loss ablation}, resulting in the following conclusions.
First, $\mathcal{L}_{NA}$ boosts performance through the query prediction refinement module and self-adaptive loss, \textit{e.g.}, TCR improves \text{R@1} by 9.2\% and 14.6\% on text and image retrieval, compared to Tent in Table~\ref{tab: both-domain}.
Second, both the designed intra-modality uniformity learning module and inter-modality gap learning module would enhance robustness against query shift.
Third, TCR achieves optimal performance when all the loss terms are employed.
Moreover, we carry out experiments to verify the effectiveness of the proposed query prediction refinement module in Section~\ref{sec: noise-robust adaptation}.
As shown in Fig.~\ref{fig: ablation}(a), we observe that: i) selecting an appropriate temperature for the existing TTA approach across various datasets is challenging; ii) even a low temperature (\textit{e.g.}, $1e-4$) is a better setting across all datasets, the performance degrades as a low temperature tends to make model overfitting on noisy query prediction.
In contrast, the query prediction refinement module not only stabilizes the temperature setting for all the datasets but also prevents the model from either underfitting or overfitting by excluding some irrelevant samples in the gallery.

\textbf{Visualization Result.}
To qualitatively study the effectiveness of TCR, we conduct the t-SNE visualization on both the query and gallery embeddings before and after the TTA process.
From the results in Fig.~\ref{fig: ablation}(c), one could observe that the samples in the query modality enjoy more scatter and the difference between the query and gallery modalities narrows after the TTA process.
In other words, TCR achieves better robustness against query shift by rectifying the intra-modality uniformity and the inter-modality gap.
\section{Conclusion}
In this paper, we develop a new test-time adaptation method, dubbed TCR, to achieve robust cross-modal retrieval against the query shift problem.
In brief, TCR first employs a novel module to refine the query predictions.
After that, TCR adopts a novel joint objective to prevent the model adaptation from overfitting noise while simultaneously manipulating the intra-modality uniformity and inter-modality gap to preserve the well-established common space from the source model.
Extensive experiments not only verify the effectiveness of TCR but also reveal the importance of each designs.
In the future, we plan to explore more potential scenarios contaminated with query shift and extend TCR to address the corresponding issues.

\bibliography{iclr2025_conference}
\bibliographystyle{iclr2025_conference}

\newpage
\appendix
\begin{leftline}
	{
		\LARGE{\textsc{Appendix}}
	}
\end{leftline}

	\etocdepthtag.toc{mtappendix}
    \etocsettagdepth{mtchapter}{none}
    \etocsettagdepth{mtappendix}{subsection}
    
    {
        \hypersetup{linkcolor=black}
    	\footnotesize\tableofcontents
    }

\newpage
\section{Defination of Intra-modality Uniformity and Inter-modality Gap}
\label{Appendix: def intra and inter}
Here, we provide a mathematical definition of the modality uniformity and the modality gap mentioned in the manuscript.
Specifically, the modality uniformity of query modality is defined as
\begin{equation}
    \text{Uniformity}=\frac{1}{N^{Q}}\sum_{i=1}^{N^{Q}}\|\mathbf{z}_{i}^{Q}-\overline{\mathbf{Z}}^{Q}\|.
    \label{eq: def uni}
\end{equation}
A low intra-modality uniformity illustrates that the samples in the query modality are compact, which degrades the retrieval performance as the model struggles to distinguish diverse queries.

The modality gap between query and gallery modalities is defined as
\begin{equation}
    \text{Modality Gap}=\|\overline{\mathbf{Z}}^{Q}-\overline{\mathbf{Z}}^{G}\|.
    \label{eq: def: gap}
\end{equation}
Modality gap has been proved to be a inherent characteristic of multimodal pre-trained models in~\citet{MindGap}, which might represent the well-aligned common space to some extent. 
As shown in Fig.~\ref{fig: ablation}(a),
either an over-low or over-high modality gap would destroy the well-constructed common space, harming the retrieval performance.

\newpage

\section{More Implementation Details}
\label{Appendix: implementation details}
\subsection{More Details about the Benchmarks}
\label{Appendix: benchmark details}
In the manuscript, we employ the QS setting and the QGS setting for evaluation. Here, we provide more detail about the benchmarks employed in the two settings.

\textbf{Query Shift.}
We construct two benchmarks with only query modality distribution shifts based on the widely-used COCO and Flickr datasets.
Specifically, 
\begin{itemize}
    \item COCO is a large-scale dataset for cross-modal retrieval and image captioning tasks. For evaluation, we conduct experiments on the COCO 2014 testing set following~\citet{BLIP}, which contains 5,000 images and 25,000 annotations, with each image associated with five corresponding text descriptions.
    \item Flickr is a cross-modal retrieval dataset collected from natural scenarios. Following~\citet{CLIP}, we employ the test set comprising 1,000 images and 5,000 annotations, where each image is paired with five corresponding sentences.
\end{itemize}
Following~\cite{MMbenchmark}, we introduce 16 and 15 types of corruption to the image and text modality, respectively.
Specifically, the corruptions in image modality consist of: 
(1) Noise: Gaussian noise, Shot noise, Impulse noise, Speckle noise; 
(2) Blur: Defocus blur, Glass blur, Motion blur, Zoom blur ; 
(3) Weather: Snow, Frost, Fog, Brightness; 
(4) Digital: Contrast, Elastic, Pixelate, JPEG compression. 
To simulate real-world corruptions, each image modality corruption is applied at five different severity levels, resulting in a total of 80 perturbations.
As for the text modality, the employed corruptions could be categorized into three levels: character-level, word-level, and sentence-level.
Specifically, the character-level corruptions consist of OCR, Character Insert (CI), Character Replace (CR), Character Swap (CS), and Character Delete (CD), which simulate real-world typos or mistakes during typing.
The word-level corruptions involve Synonym Replacement (SR), Word Insertion (WR), Word Swap (WS), Word Deletion (WD), and Insert Punctuation (IP), which simulate different writing habits that people may replace, delete, or add words to express the same meaning.
For sentence-level corruptions, we convert the annotation styles into Formal, Casual, Passive, Active, and Back-translation, which simulate various speaking, writing styles or translation errors. 
Similar to the image corruptions, we introduce 7/2/1 severity levels for character-level/word-level/sentence-level corruptions.

\begin{figure*}[t]
\centering
\vspace{-0.15in}
\includegraphics[width=0.9\linewidth]{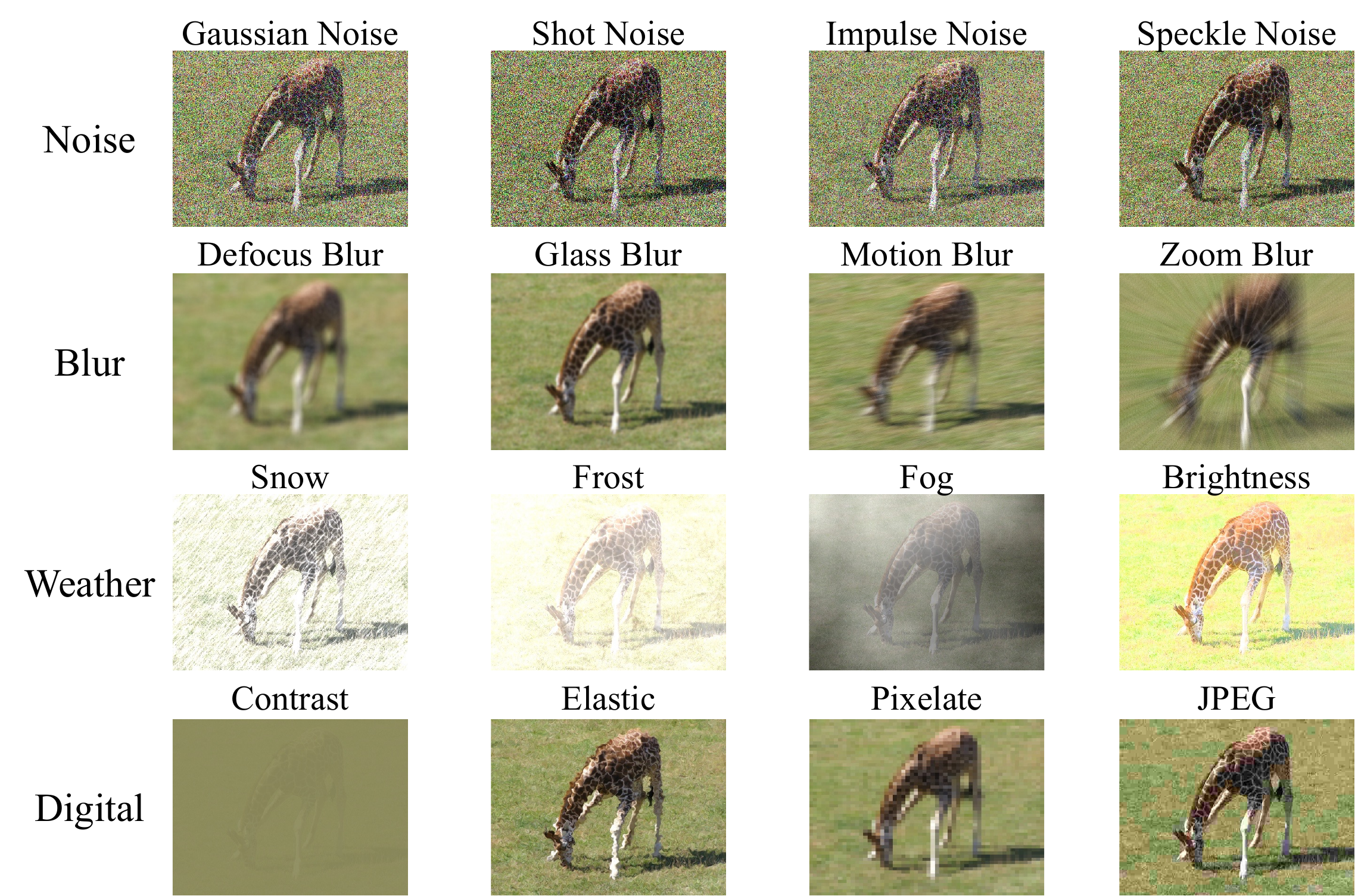}
\vspace{-0.1in}
\caption{Examples of 16 types of image corruption. The original image is from the COCO dataset.}
\vspace{-0.1in}
\label{fig: image-corruption-example}
\end{figure*}

As a result, we construct the two benchmarks named COCO-C and Flickr-C.
Notably, we only introduce the corruptions to the query modality in the QS setting, \textit{e.g.}, for image-to-text retrieval, the distribution shifts occur on the image modality.
The cases of the 16 image corruptions and 15 text corruptions are visualized in Fig.~\ref{fig: image-corruption-example} and Table~\ref{tab: text-corruption-example}.

\textbf{Query-Gallery Shift.} 
We establish the following benchmarks with distribution shifts across both query and gallery modalities, including Fashion-Gen from the E-commerce domain, CUHK-PEDES and ICFG-PEDES from the ReID domain, as well as COCO, Flickr, and Nocaps from the natural image domain.
Specifically,
\begin{itemize}
    \item Fashion-Gen is a cross-modal retrieval dataset source from the E-commerce domain, comprising fashion images paired with item descriptions provided by professional stylists. In the experiment, we employ the testing set containing 32,528 image-text pairs. 
    \item CUHK-PEDES is a dataset for text-to-image person re-identification. Following~\citep{IRRA}, we utilize the testing set which contains 3,074 images and 6,156 textual descriptions of 1,000 identities.
    \item ICFG-PEDES is a large-scale text-to-image person re-identification dataset. Following~\citep{IRRA}, we adopt the testing set consists of 19,848 image-text pairs, corresponding to 1,000 identities.
    \item Nocaps is a cross-modal retrieval dataset derived from the OpenImages dataset. For evaluation, we perform experiments on the test set, which consists of 648 in-domain images, 2,938 near-domain images, and 914 out-domain images. Each image is paired with 10 captions.
\end{itemize}
\begin{table*}[t]
    \vspace{-0.1in}
    \caption{Examples of 15 types of text corruption. The original text is from the COCO dataset.}
    \vspace{-0.05in}
    \label{tab: text-corruption-example}
\newcommand{\tabincell}[2]{\begin{tabular}{@{}#1@{}}#2\end{tabular}}
 \begin{center}
 \begin{threeparttable}
 \small
    \resizebox{0.9\linewidth}{!}{
        \begin{tabular}{ll|p{9.5cm}}
            \toprule
              Category & Perturbation  & Example   \\ 
            \midrule
            \multirow{1}{*}{Original } 
             &Clean  & A train traveling down tracks next to a brick building.	 \\
             \midrule
             \multirow{7}{*}{Character} 
             &OCR  & A train travelin\textcolor{orange}{9} down track\textcolor{orange}{8} next to a brick building. \\ 
             \cmidrule(l){2-2} \cmidrule(l){3-3} 
             &CI  & A train traveling down tra\textcolor{orange}{G}cks next to a brick bui\textcolor{orange}{1}lding. \\
             \cmidrule(l){2-2} \cmidrule(l){3-3} 
             &CR & A train traveling do\textcolor{orange}{P}n tracks next to a brick buildi\textcolor{orange}{r}g.  \\
             \cmidrule(l){2-2} \cmidrule(l){3-3} 
             &CS  & A train r\textcolor{orange}{t}ave\textcolor{orange}{i}\textcolor{orange}{l}ng down tracks next to a brick building. \\ 
             \cmidrule(l){2-2} \cmidrule(l){3-3} 
             &CD  & A train tr\textcolor{orange}{[X]}veling down tr\textcolor{orange}{[X]}cks next to a brick building. \\
             \midrule
             \multirow{6}{*}{Word}
             &SR  & A train \textcolor{orange}{jaunt} down running \textcolor{orange}{adjacent} to a brick building.
        	 \\  
             \cmidrule(l){2-2} \cmidrule(l){3-3} 
             &RI  & A train \textcolor{orange}{pass} traveling down tracks next to \textcolor{orange}{go} a brick building
        	 \\ 
             \cmidrule(l){2-2} \cmidrule(l){3-3} 
             &RS  & A \textcolor{orange}{building} traveling down tracks next to a brick \textcolor{orange}{train}.
        	 \\ 
             \cmidrule(l){2-2} \cmidrule(l){3-3} 
             &RD  & A train \textcolor{orange}{[X]} down tracks \textcolor{orange}{[X]} to a brick building.
        	 \\ 
             \cmidrule(l){2-2} \cmidrule(l){3-3}
             &IP  & A \textcolor{orange}{:} train traveling down tracks next to \textcolor{orange}{,} a brick building. \\
             \midrule
             \multirow{8}{*}{Sentence} 
             &Formal & A train \textcolor{orange}{moving} down tracks next to a brick building.	  \\
             \cmidrule(l){2-2} \cmidrule(l){3-3} 
             &Casual  & A train \textcolor{orange}{that goes down tracks} next to a brick building.	  \\
             \cmidrule(l){2-2} \cmidrule(l){3-3} 
             &Passive 	& Tracks next to a brick building \textcolor{orange}{are being traveled down by} a train. \\
             \cmidrule(l){2-2} \cmidrule(l){3-3} 
             &Active  & \textcolor{orange}{There is} a train traveling down tracks next to a brick building.  \\
             \cmidrule(l){2-2} \cmidrule(l){3-3} 
             &Backtrans   & A train \textcolor{orange}{runs down the} tracks next to a brick building. \\
            \bottomrule
        \end{tabular}
        }
	 \end{threeparttable}
	 \end{center}
\vspace{-0.3in}
\end{table*}
\subsection{More Experiment Details}
\label{Appendix: experiment details}
To guarantee the performance of the baselines, we select the optimal temperature (Eq.~\ref{eq: prediction}) for the TTA baselines upon each dataset.
According to Fig.~\ref{fig: ablation}(a), the temperature is fixed as $0.01$ for COCO-C, Flickr-C, COCO, Flickr, and Nocaps datatsets, $0.001$ for Fashion-Gen dataset, and $0.0001$ for CUHK-PEDE and ICFG-PEDES datasets.
Note that we employ Tent to conduct the experiment in Fig.~\ref{fig: ablation}(a) since most TTA methods are variants based on the Tent.
Moreover, the adaptation process utilizes an initial learning rate of $3e^{-4}/3e^{-5}$ for text/image retrieval, excepting $3e^{-4}$ for image retrieval on the CLIP model.


In addition, for the ablation study in Fig.~\ref{fig: ablation}(a), we perform experiments on the Flickr-C dataset using the following corruptions: Gaussian, Zoom, Snow, and Contrast for the image modality; OCR, IP, and Formal for the text modality.
\subsection{Pseudo Code}
\label{Appendix: pseudo code}
In the following, we provide the pseudo-code of the proposed TCR in Algorithm~\ref{alg: tcr-algorithm}. 
To guarantee the stability of the estimation for $E_m$ and $\Delta_{S}$, we maintain a queue which always saves the query-candidate pairs with the smallest SI during the adaptation process.
Following ~\cite{SWAV}, we limit the queue updating times to a maximum of 10 iterations.
\begin{figure}[h]
    \vspace{-0in}
    \centering
    \begin{minipage}{1.0\textwidth}
        \centering
        \begin{algorithm}[H]\small
         \KwIn{Test samples $\mathcal{D}_{T} = \left\{\{\mathbf{x}_{i}^{Q}\}_{i=1}^{N^{Q}}, \{\mathbf{x}_{j}^{G}\}_{j=1}^{N^{G}}\right\}$, the source model $f_{\Theta_{s}}$ with trainable parameters $\tilde{\Theta}$, TTA steps $\eta>0$, batch size $B$.}
         \KwOut{Predictions $\{\mathbf{p}_i\}_{i=1}^{N^{Q}}$.}
         Initialize $\tilde{\Theta}_0=\Theta_{s}$\;
         \For{given queries $\mathbf{x}^{Q} \in \mathcal{D}_{T}$}{
         \For{$\text{step}=1,\cdots,\eta$}{
         Obtain the query predictions $\mathbf{p}$ in Eq.~\ref{eq: prediction}\;
         Select a subset of candidates $\mathbf{x}^{G^{\prime}}$ from the gallery using Eq.~\ref{eq: construct gallery}\ \tcp*{Candidate Selection}
         Obtain the refined query predictions $\hat{\mathbf{p}}$ in Eq.~\ref{eq: new prediction} and the corresponding entropy $E(\mathbf{x}^{Q})$\;
          \tcp{Update the queue}
          \If{$step=1$}{ 
                Compute the criterion $\text{SI}$ in Eq.~\ref{eq: SI}\;
                Select the $30\%$ query-candidate pairs with the smallest $SI$\;
                Maintain a queue of size $B$ to save the pairs and their corresponding entropies\;
          }
          Estimate the modality gap $\Delta_{S}$ using Eq.~\ref{eq: estimate mg}\ \tcp*{Constraint Estimation}
          Estimate the desirable threshold $E_{m}$ using Eq.~\ref{eq: get_max_entropy}\ \tcp*{Constraint Estimation}
          Compute the overall loss $\mathcal{L}$ in Eq.~\ref{eq: overall} with $\Delta_{S}$ and $E_{m}$\; 
          Update parameters $\tilde{\Theta}$ through gradient descent to minimize $\mathcal{L}$;
          }
         }
         \caption{\textbf{T}est-time adaptation for \textbf{C}ross-modal \textbf{R}etrieval (\methodname)}
         \label{alg: tcr-algorithm}
        \end{algorithm}
    \end{minipage}%
    \vspace{-0.4in}
\end{figure}

\newpage
\section{An Alternative Implementation of TCR without training}
\label{Appendix: untrained TCR}
From the results in Fig.~\ref{fig: analytic-two-charac}, we observe that the performance could be boosted by simply scaling up the uniformity or rectifying the modality gap even without adopting the function.
Based on the observation, in this section, we propose to implement TCR in an untrained manner, which demonstrates the great potential of TCR.
Specifically, to enhance the uniformity of the query modality, 
we scale the given queries by Eq.~\ref{eq: scale} with $\lambda^{\text{scale}}$ fixed at $2$.
To adjust the inter-modality gap, we estimate the modality gap $\Delta_{S}$ of the source domain by Eq.~\ref{eq: estimate mg} and then rectify the modality gap in the target domain to $\Delta_{S}$ by Eq.~\ref{eq: offset}.
The details are presented in Algorithm~\ref{alg: tcr-untrain-algorithm} and the experiment results are shown in Table~\ref{tab: untrain-tcr-image-c}-\ref{tab: untrain-tcr-text-c}.
\begin{figure}[h]
    \centering
    \begin{minipage}{1.0\textwidth}
        \centering
        \begin{algorithm}[H]\small
         \KwIn{Test samples $\mathcal{D}_{T} = \left\{\{\mathbf{x}_{i}^{Q}\}_{i=1}^{N^{Q}}, \{\mathbf{x}_{j}^{G}\}_{j=1}^{N^{G}}\right\}$, the source model $f_{\Theta_{s}}$, batch size $B$, scaling factor $\lambda^{\text{scale}}$.}
         \KwOut{Predictions $\{\mathbf{p}_i\}_{i=1}^{N^{Q}}$.}
         Initialize $\tilde{\Theta}_0=\Theta_{s}$\;
         \For{given queries $\mathbf{x}^{Q} \in \mathcal{D}_{T}$}{
         Select a subset of candidates $\mathbf{x}^{G^{\prime}}$ from the gallery using Eq.~\ref{eq: construct gallery}\ \tcp*{Candidate Selection}
          \tcp{Update the queue}
            Compute the criterion $\text{SI}$ in Eq.~\ref{eq: SI}\;
            Select the $30\%$ query-candidate pairs with the smallest $SI$\;
            Maintain a queue of size $B$ to save the pairs\;
          Scale $\mathbf{x}^{Q}$ using Eq.~\ref{eq: scale} with $\lambda^{\text{scale}}$\ \tcp*{Scaling up Intra-modality Uniformity}
          Estimate the modality gap $\Delta_{S}$ using Eq.~\ref{eq: estimate mg}\ \tcp*{Constraint Estimation}
          Rectify the modality gap to $\Delta_{S}$ using Eq.~\ref{eq: offset}\ \tcp*{Rectifying between-modality Gap}
          Perform $\ell 2$-normalization on the embeddings in the query modality\;
          Obtain the query predictions $\mathbf{p}$ in Eq.~\ref{eq: prediction};
         }
         \caption{An Implementation of Untrained TCR}
         \label{alg: tcr-untrain-algorithm}
        \end{algorithm}
    \end{minipage}%
\end{figure}

\newpage

\begin{table*}[t]
    \vspace{-0.1in}
    \caption{
    Comparisons with state-of-the-art methods on Flickr-C benchmark under \textbf{\textsc{query shift on the image modality}} with maximum severity level regarding the Recall@1 metric. 
    }
    \vspace{-0.05in}
    \label{tab:flickr-c-image-domain-shift}
\newcommand{\tabincell}[2]{\begin{tabular}{@{}#1@{}}#2\end{tabular}}
 \begin{center}
 \begin{threeparttable}
 \LARGE
    \resizebox{0.9\linewidth}{!}{
 	\begin{tabular}{l|cccc|cccc|cccc|cccc|>{\columncolor{blue!8}}c}
 	\multicolumn{1}{c}{} & \multicolumn{4}{c}{Noise} & \multicolumn{4}{c}{Blur} & \multicolumn{4}{c}{Weather} & \multicolumn{4}{c}{Digital}  \\
 	Query Shift & Gauss. & Shot & Impul. &Speckle & Defoc. & Glass & Motion & Zoom & Snow & Frost & Fog & Brit. & Contr. & Elastic & Pixel & JPEG & Avg.  \\
    \cmidrule{1-18}
        BLIP ViT-B/16 &  49.8 & 56.6 &  50.3 & 71.6 & 53.1 & 84.5 & 47.4 & 15.5 & 66.4 & 80.4 & 79.5 &   85.5 & 60.6 & 53.3 & 35.1 & 80.3 & 60.6\\ 
        ~~$\bullet~$Tent & 54.9 & 54.9 & 54.3 & 73.1 & 53.3 & 85.3 & 47.9 & 1.6 & 67.2 & 80.9 & 79.6 & 86.8 & 63.6 & 53.4 & 35.4 & 81.4 & 60.9  \\ %
        ~~$\bullet~$EATA & 55.5 & 60.5 & 55.8 & 75.8 & 64.6 & 86.2 & 52.2 & 8.5  & 72.0 & 83.7 & 82.5 & 87.9 & 68.4 & 60.1 & 45.9 & 81.6 & 65.1 \\
        ~~$\bullet~$SAR & 54.8 & 62.5 & 55.6 & 75.2 & 48.3 & 87.2 & 34.8 & 15.5 & 71.9 & 83.1 & 82.2 & 87.9 & 68.2 & 60.3 & 42.2 & 81.4 & 63.2  \\
        ~~$\bullet~$READ & 50.1 & 58.2 & 52.2 & 74.8 & 63.7 & 87.0 & 55.1 & 2.2  & 71.7 & 83.8 & 81.9 & 87.7 & 67.4 & 62.3 & 42.5 & 81.4 & 63.9  \\
        ~~$\bullet~$DeYO & 55.4 & 62.0 & 56.3 & 76.2 & 63.8 & 86.3 & 50.3 & 3.2 & 73.1 & 84.1 & 83.2 & 88.6 & 70.1 & 63.1 & 46.8 & 81.3 & 65.2  \\
        \rowcolor{pink!30}~~$\bullet~$Ours & \textbf{62.0} & \textbf{66.6} & \textbf{61.4} & \textbf{80.0} & \textbf{68.1} & \textbf{87.9} & \textbf{65.2} & \textbf{39.9} & \textbf{78.2} & \textbf{85.2} & \textbf{85.7} & \textbf{89.5} & \textbf{75.1} & \textbf{73.1} & \textbf{56.8} & \textbf{83.3} & \textbf{72.4} \\
    \cmidrule{1-18}
        BLIP ViT-L/16 &  58.2 & 61.0 &  59.7 & 76.9 & 66.4 & 88.5 & 62.5 & 33.4 & 67.7 & 81.5 & 79.3 & 89.1 & 60.4 & 66.4 & 46.5 & 85.0 & 67.7\\ 
        ~~$\bullet~$Tent & 61.3 & 64.3 & 63.3 & 77.6 & 70.8 & 88.7 & 62.8 & 31.5 & 70.4 & 83.8 & 81.1 & 89.2 & 61.2 & 68.7 & 52.0   & 84.5 & 69.5 \\ %
        ~~$\bullet~$EATA & 62.0 & 65.1 & 64.5 & 78.9 & 70.2 & 89.5 & 63.3 & 33.1 & 71.9 & 83.7 & 81.2 & 89.3 & 61.6 & 69.3 & 53.0 & 85.8 & 70.2  \\
        ~~$\bullet~$SAR & 61.1 & 64.4 & 63.7 & 79.7 & 71.6 & 90.3 & 64.4 & 27.6 & 70.6 & 83.4 & 81.0 & 89.7 & 62.4 & 70.1 & 53.3 & 85.3 & 69.9  \\
        ~~$\bullet~$READ & 61.1 & 64.4 & 63.7 & 79.7 & 71.6 & 90.3 & 64.4 & 27.6 & 70.6 & 83.4 & 81.0 & 89.7 & 62.4 & 70.1 & 53.3 & 85.3 & 69.9  \\
        ~~$\bullet~$DeYO & 61.5 & 61.0 & 62.1 & 78.3 & 69.6 & 89.5 & 62.5 & 37.2 & 72.1 & 83.6 & 81.4 & 89.9 & 61.3 & 67.6 & 52.5 & 86.8 & 69.8 \\
        \rowcolor{pink!30}~~$\bullet~$Ours & \textbf{68.2} & \textbf{71.7} & \textbf{70.2} & \textbf{83.3} & \textbf{74.7} & \textbf{91.9} & \textbf{72.5} & \textbf{49.6} & \textbf{78.2} & \textbf{87.0} & \textbf{85.5} & \textbf{92.1} & \textbf{70.9} & \textbf{79.6} & \textbf{65.5} & \textbf{87.8} & \textbf{76.8} \\
    \cmidrule{1-18}
	\end{tabular}
	}
	 \end{threeparttable}
	 \end{center}
\vspace{-0.2in}
\end{table*}

\begin{table*}[t]
    \caption{Comparisons with state-of-the-art methods on Flickr-C benchmark under \textbf{\textsc{query shift on the text modality}} with maximum severity level regarding the Recall@1 metric. 
    }
    \vspace{-0.05in}
    \label{tab:flickr-c-text-domain-shift}
\newcommand{\tabincell}[2]{\begin{tabular}{@{}#1@{}}#2\end{tabular}}
 \begin{center}
 \begin{threeparttable}
 \LARGE
    \resizebox{0.9\linewidth}{!}{
 	\begin{tabular}{l|ccccc|ccccc|ccccc|>{\columncolor{blue!8}}c}
 	\multicolumn{1}{c}{} & \multicolumn{5}{c}{Character-level} & \multicolumn{5}{c}{Word-level} & \multicolumn{5}{c}{Setence-level}  \\
 	 Query Shift & OCR & CI & CR & CS & CD & SR & RI & RS & RD & IP & Formal & Casual & Passive & Active & Backtrans & Avg.  \\
    \cmidrule{1-17}
        BLIP ViT-B/16 & 53.5 & 18.4 & 18.0 & 30.4 & 22.5 & 68.3 & 77.9 & 76.9 & 77.9 & 82.1 & 82.1 & 81.9 & 79.9 & 82.2 & 79.8 & 62.1 \\ 
        ~~$\bullet~$Tent & 55.4 & 18.6 & 18.2 & 31.1 & 23.0 & 69.6 & 78.8 & 77.7 & 78.0 & 82.2 & 81.9 & 81.8 & 79.6 & 82.0 & 79.9 & 62.5    \\ %
        ~~$\bullet~$EATA & 55.7 & 19.9 & 19.9 & 31.6 & 23.6 & 69.5 & 78.6 & 77.5 & 77.9 & 82.4 & 82.3 & 81.8 & 80.5 & \textbf{82.6} & 80.2 & 62.9    \\
        ~~$\bullet~$SAR & 53.5 & 20.1 & 19.1 & 32.1 & 23.8 & 68.3 & 77.9 & 76.9 & 77.9 & 82.1 & 82.1 & 81.9 & 79.9 & 82.2 & 79.8 & 62.5    \\
        ~~$\bullet~$READ & 55.8 & 19.7 & 20.6 & 32.0 & 23.5 & 69.3 & 78.6 & 77.6 & 78.1 & 82.4 & 82.2 & 81.8 & 80.5 & 82.5 & 80.2 & 63.0    \\
        ~~$\bullet~$DeYO & 53.5 & 18.4 & 18.0 & 30.4 & 22.5 & 68.3 & 77.9 & 76.9 & 77.9 & 82.1 & 82.1 & 81.9 & 79.9 & 82.2 & 79.8 & 62.1    \\
        \rowcolor{pink!30}~~$\bullet~$Ours & \textbf{57.1} & \textbf{21.4} & \textbf{22.5} & \textbf{33.6} & \textbf{25.1} & \textbf{69.8} & \textbf{79.3} & \textbf{78.0} & \textbf{78.1} & \textbf{82.5} & \textbf{82.4} & \textbf{82.2} & \textbf{81.0} & \textbf{82.6} & \textbf{80.2} & \textbf{63.7} \\

    \cmidrule{1-17}
        BLIP ViT-L/16 &58.0 & 22.2 & 22.0 & 34.1 & 25.1 & 71.2 & 79.9 & 78.9 & 78.8 & 83.3 & 83.1 & 82.7 & 81.7 & 83.5 & 80.7 & 64.4 \\
        ~~$\bullet~$Tent & 59.0 & 22.4 & 22.1 & 34.5 & 25.3 & 71.4 & 80.3 & 79.3 & 78.8 & 83.7 & 82.8 & 82.7 & 81.8 & 83.3 & 80.7 & 64.6 \\
        ~~$\bullet~$EATA & 59.1 & 23.0 & 23.2 & 35.1 & 25.6 & 71.7 & 80.3 & 79.3 & 78.8 & 83.5 & 83.0 & 83.2 & 81.8 & \textbf{83.5} & 80.7 & 64.8  \\
        ~~$\bullet~$SAR & 58.1 & 23.1 & 23.0 & 34.5 & 25.8 & 71.2 & 79.9 & 78.9 & 78.8 & 83.3 & 83.1 & 82.7 & 81.7 & 83.4 & 80.7 & 64.6  \\
        ~~$\bullet~$READ & 58.9 & 23.4 & 23.3 & 34.9 & 25.9 & 71.5 & 80.7 & 79.3 & 78.8 & 83.5 & \textbf{83.2} & 83.1 & \textbf{81.9} & 83.4 & \textbf{80.8} & 64.8  \\
        ~~$\bullet~$DeYO & 58.1 & 22.2 & 22.0 & 34.1 & 25.1 & 71.2 & 79.9 & 78.9 & 78.7 & 83.3 & 83.1 & 82.7 & 81.7 & 83.4 & \textbf{80.8} & 64.4  \\
        \rowcolor{pink!30}~~$\bullet~$Ours & \textbf{59.7} & \textbf{24.4} & \textbf{24.4} & \textbf{36.1} & \textbf{26.7} & \textbf{71.8} & \textbf{80.9} & \textbf{79.5} & \textbf{78.9} & \textbf{83.5} & \textbf{83.2} & \textbf{83.4} & 81.8 & \textbf{83.5} & \textbf{80.8} & \textbf{65.2} \\
    \cmidrule{1-17}
	\end{tabular}
	}
	 \end{threeparttable}
	 \end{center}
\vspace{-0.2in}
\end{table*}

\section{More Experiment Results}
\subsection{Results on Flickr-C}
\label{Appendix: Flickr-C}
In the manuscript, we have carried out experiments on the COCO-C benchmark. Here, we provide more results on the Flickr-C benchmark.
As shown in Table~\ref{tab:flickr-c-image-domain-shift}-\ref{tab:flickr-c-text-domain-shift}, TCR significantly outperforms all the baselines across various pre-trained model types and sizes on the Flickr-C benchmarks.
\subsection{Experiment about Personalized Queries on Fashion-Gen}
\label{Appendix: Fashion-Gen details}
As mentioned in Introduction of the manuscript, different inquirers would submit personalized queries, \textit{e.g.}, some are drawn to fashionable handbags, while others are passionate about collecting a variety of shoes.
To further demonstrate the generalization of TCR in this scenario, we simulate the personalized queries on the Fashion-Gen benchmark.
In detail, following~\citet{Openfashionclip}, we fine-tune the pre-trained CLIP on four publicly available fashion datasets including Fashion-Gen, Fashion IQ~\citep{FashionIQ}, Fashion200K~\citep{Fashion200K}, and iMaterialist~\citep{iMaterialist}.
After that, we employ TCR to adapt various preferences such as ``TOPS'' and ``SWEATERS'' on the Fashion-Gen benchmark.
The experimental results are presented in Table~\ref{tab:fashion-gen}, one could observe that TCR improves both image-to-text and text-to-image retrieval performance under personalized queries.

\begin{table*}[t]
    \vspace{-0.1in}
    \caption{The cross-modal retrieval performance of TCR on Fashion-Gen benchmark with \textbf{\textsc{personalized queries }}regarding Recall@1 metric.
    }
    \vspace{-0.05in}
    \label{tab:fashion-gen}
\newcommand{\tabincell}[2]{\begin{tabular}{@{}#1@{}}#2\end{tabular}}
 \begin{center}
 \begin{threeparttable}
 \LARGE
    \resizebox{0.95\linewidth}{!}{
    \begin{tabular}{c|l|cccccccccc>{\columncolor{blue!8}}c}
                     & Query Shift & TOPS  & SWEATERS & JACKETS & PANTS & JEANS & SHIRTS & DRESSES & SHORTS & SNEAKERS & SKIRTS & Avg.   \\
                    \cmidrule{1-13}
\multirow{2}{*}{TR} & BLIP ViT-B/32 & 18.0 & 19.3 & 19.9 & 12.0 & 5.5 & 18.3 & 38.1 & \textbf{17.9} & 37.3 & 29.6 & 21.6 \\
                    & $\bullet~$Ours & \textbf{22.9} & \textbf{25.2} & \textbf{21.6} & \textbf{14.3} & \textbf{6.0} & \textbf{22.}8 & \textbf{44.3} & 8.5  & \textbf{41.7} & \textbf{37.4} & \textbf{24.5} \\
                    \cmidrule{1-13}
\multirow{2}{*}{IR} & BLIP ViT-B/32 & 24.9 & 27.9 & 29.2 & 16.9 & 6.7 & 25.4 & 51.8 & 25.7 & 47.1 & 47.8 & 30.3 \\
                    & $\bullet~$Ours & \textbf{28.2} & \textbf{31.7} & \textbf{32.8} & \textbf{19.5} & \textbf{9.6} & \textbf{28.5} & \textbf{57.1} & \textbf{29.1} & \textbf{53.6} & \textbf{50.7} & \textbf{34.1} \\
                    \cmidrule{1-13}
\end{tabular}}
	 \end{threeparttable}
	 \end{center}
\vspace{-0.2in}
\end{table*}
\begin{figure*}[t]
\centering
\includegraphics[width=0.9\linewidth]{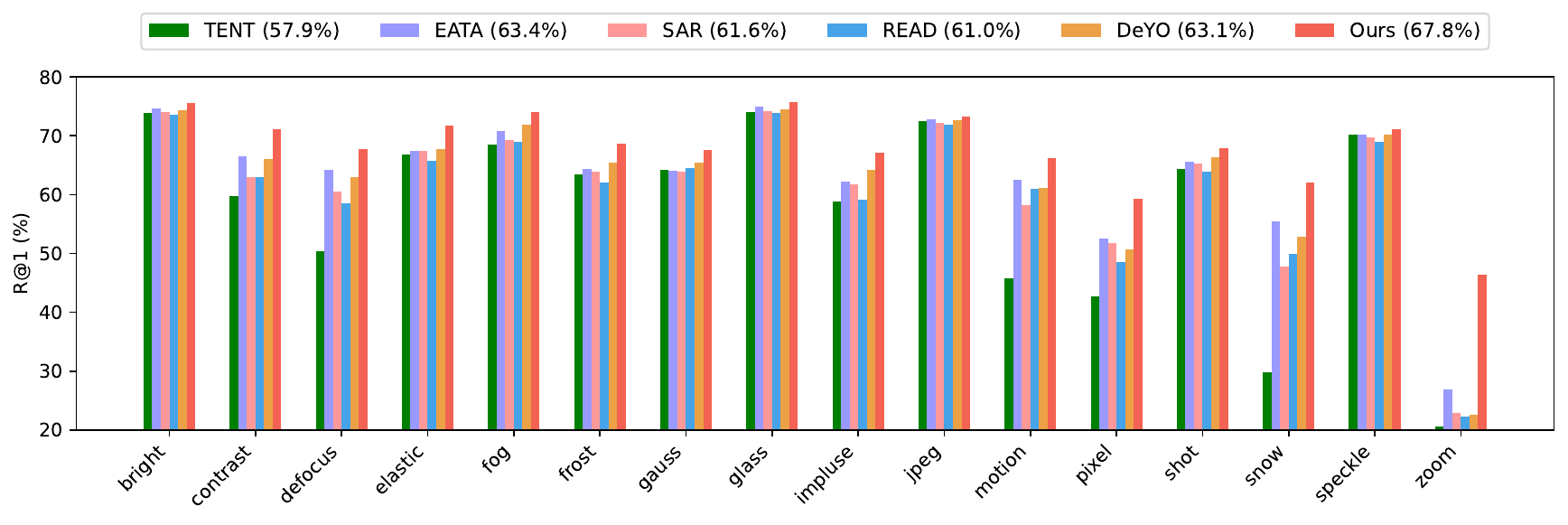}
\vspace{-0.2in}
\caption{Text retrieval performance comparisons on the COCO-C benchmark under \textbf{\textsc{query shift on the image modality}} with all severity levels regarding Recall@1 metric. The legend key provides an overview of the average performance of each approach across various corruption types.
}
\label{fig: all_i2t}
\end{figure*}
\begin{figure*}[!t]
\centering
\includegraphics[width=0.9\linewidth]{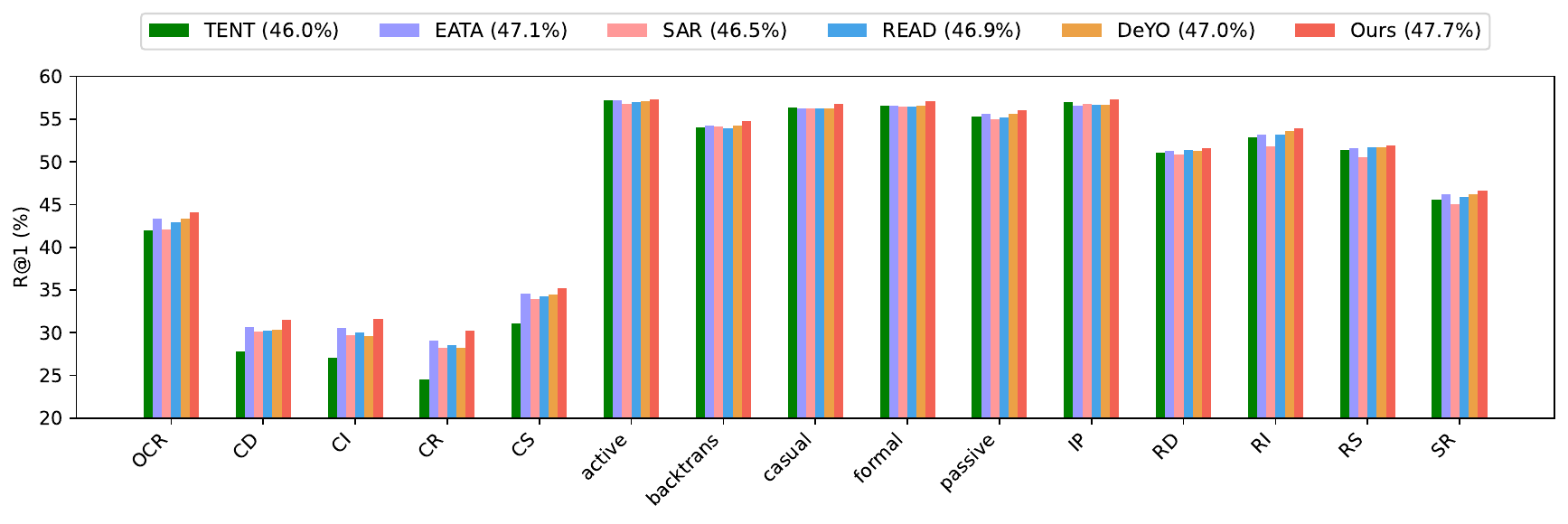}
\vspace{-0.2in}
\caption{Image retrieval performance comparisons on the COCO-C benchmark under \textbf{\textsc{query shift on the text modality}} with all severity levels regarding Recall@1 metric. }
\label{fig: all_t2i}
\end{figure*}
\subsection{Results on the All Severity Setting}
\label{Appendix: all severity}
In Section~\ref{sec: comparisons SOTA} of the manuscript and Appendix~\ref{Appendix: Flickr-C}, we have demonstrated the effectiveness of TCR in handling query shift at the maximum severity level.
To further verify the robustness of TCR, we conduct more experiments on the COCO-C benchmark with query shift across all severity levels.

The results in Fig.~\ref{fig: all_i2t}-\ref{fig: all_t2i} indicate the effectiveness of TCR in addressing various severity of the query shift.
\begin{table*}[t]
    \vspace{-0.1in}
    \caption{The cross-modal retrieval performance of untrained TCR on COCO-C and Flickr-C benchmarks under \textbf{\textsc{image modality distribution shifts}} with maximum severity level regarding the Recall@1 metric.
    }
    \vspace{-0.05in}
    \label{tab: untrain-tcr-image-c}
\newcommand{\tabincell}[2]{\begin{tabular}{@{}#1@{}}#2\end{tabular}}
 \begin{center}
 \begin{threeparttable}
 \LARGE
    \resizebox{0.95\linewidth}{!}{
 	\begin{tabular}{c|l|cccc|cccc|cccc|cccc|>{\columncolor{blue!8}}c}
 	\multicolumn{1}{c}{} & \multicolumn{1}{c}{} & \multicolumn{4}{c}{Noise} & \multicolumn{4}{c}{Blur} & \multicolumn{4}{c}{Weather} & \multicolumn{4}{c}{Digital}  \\
 	Dataset & Query Shift & Gauss. & Shot & Impul. &Speckle & Defoc. & Glass & Motion & Zoom & Snow & Frost & Fog & Brit. & Contr. & Elastic & Pixel & JPEG & Avg.  \\
    \cmidrule{1-19}
        \multirow{4}{*}{Flickr-C} & EATA & 55.5 & 60.5 & 55.8 & 75.8 & 64.6 & 86.2 & 52.2 & 8.5  & 72.0 & 83.7 & 82.5 & 87.9 & 68.4 & 60.1 & 45.9 & 81.6 & 65.1 \\
        & Ours (untrain) & 58.7 & 63.2 & 58.1 & 78.8 & 65.9 & 87.8 & 61.2 & 34.6 & 79.2 & 84.8 & 84.4 & 89.1 & 68.2 & 67.4 & 46.0 & 83.0 & 69.4 \\
        & Ours & \textbf{62.0} & \textbf{66.6} & \textbf{61.4} & \textbf{80.0} & \textbf{68.1} & \textbf{87.9} & \textbf{65.2} & \textbf{39.9} & \textbf{78.2} & \textbf{85.2} & \textbf{85.7} & \textbf{89.5} & \textbf{75.1} & \textbf{73.1} & \textbf{56.8} & \textbf{83.3} & \textbf{72.4} \\
     \cmidrule{1-19}
        \multirow{4}{*}{COCO-C} & EATA & 41.4 & 50.3 & 35.7 & 63.1 & 49.8 & 72.2 & 46.2 & 6.9  & 45.6 & 56.7 & 62.5 & 71.4 & 43.6 & 51.3 & 25.6 & 67.0 & 49.3 \\
        & Ours (untrain) & 48.8 & 51.7 & 49.8 & 61.5 & 53.9 & 72.6 & 49.4 & 18.7 & 49.7 & 60.5 & 67.1 & 71.4 & 43.9 & 49.9 & 26.7 & 67.4 & 52.7  \\
        & Ours & \textbf{53.2} & \textbf{56.2} & \textbf{54.8} & \textbf{64.6} & \textbf{58.0} & \textbf{73.7} & \textbf{56.4} & \textbf{32.2} & \textbf{56.5} & \textbf{64.1} & \textbf{71.0} & \textbf{73.4} & \textbf{57.9} & \textbf{63.7} & \textbf{41.8} & \textbf{68.4} & \textbf{59.1} \\
    \cmidrule{1-19}
	\end{tabular}
	}
	 \end{threeparttable}
	 \end{center}
\vspace{-0.2in}
\end{table*}

\begin{table*}[t]
    \vspace{-0.1in}
    \caption{The cross-modal retrieval performance of untrained TCR on COCO-C and Flickr-C benchmarks under \textbf{\textsc{text modality distribution shifts}} with maximum severity level regarding the Recall@1 metric.
    }
    \vspace{-0.05in}
    \label{tab: untrain-tcr-text-c}
\newcommand{\tabincell}[2]{\begin{tabular}{@{}#1@{}}#2\end{tabular}}
 \begin{center}
 \begin{threeparttable}
 \LARGE
    \resizebox{0.95\linewidth}{!}{
 	\begin{tabular}{c|l|ccccc|ccccc|ccccc|>{\columncolor{blue!8}}c}
 	\multicolumn{1}{c}{} & \multicolumn{1}{c}{} & \multicolumn{5}{c}{Character-level} & \multicolumn{5}{c}{Word-level} & \multicolumn{5}{c}{Setence-level}  \\
 	Dataset & Query Shift & OCR & CI & CR & CS & CD & SR & RI & RS & RD & IP & Formal & Casual & Passive & Active & Backtrans & Avg.  \\
    \cmidrule{1-18}
        \multirow{3}{*}{Flickr-C} & EATA & 55.7 & 19.9 & 19.9 & 31.6 & 23.6 & 69.5 & 78.6 & 77.5 & 77.9 & 82.4 & 82.3 & 81.8 & 80.5 & \textbf{82.6} & 80.2 & 62.9    \\
        & Ours (untrain) & 55.8 & 20.3 & 20.7 & 32.7 & 23.8 & 69.2 & 78.3 & 77.8 & 77.8 & 82.5 & 82.2 & 82.0 & 80.4 & 82.3 & 80.0 & 63.1\\
        & Ours & \textbf{57.1} & \textbf{21.4} & \textbf{22.5} & \textbf{33.6} & \textbf{25.1} & \textbf{69.8} & \textbf{79.3} & \textbf{78.0} & \textbf{78.1} & \textbf{82.5} & \textbf{82.4} & \textbf{82.2} & \textbf{81.0} & \textbf{82.6} & \textbf{80.2} & \textbf{63.7} \\
    \cmidrule{1-18}
        \multirow{3}{*}{COCO-C} & EATA & 33.1 & 11.9 & 10.5 & 18.4 & 12.0 & 44.9 & 53.0 & 51.6 & 50.3 & 56.2 & 56.8 & 56.8 & 56.0 & 56.8 & 54.3 & 41.5  \\
        & Ours (untrain) & 32.9 & 12.3 & 10.4 & 19.0 & 12.3 & 44.8 & 52.6 & 51.3 & 51.5 & 57.8 & 57.1 & \textbf{57.2} & \textbf{56.2} & 57.2 & \textbf{54.7} & 41.8\\
        & Ours & \textbf{34.1} & \textbf{13.7} & \textbf{11.8} & \textbf{19.5} & \textbf{13.2} & \textbf{45.3} & \textbf{53.8} & \textbf{51.8} & \textbf{51.5} & \textbf{57.3} & \textbf{57.1} & 56.8 & 56.0 & \textbf{57.3} & \textbf{54.7} & \textbf{42.3} \\
    \cmidrule{1-18}
	\end{tabular}
	}
	 \end{threeparttable}
	 \end{center}
\vspace{-0.2in}
\end{table*}

\subsection{Results of Untrained TCR}
\label{Appendix: experiments of untrained TCR}
Here, we perform experiments on the COCO-C and Flickr-C benchmarks to evaluate the proposed untrained TCR in Appendix~\ref{Appendix: untrained TCR}.
During the experiments, we compare untrained TCR with the best baseline EATA in Table~\ref{tab: coco-c-image}-\ref{tab: coco-c-text}.
From the results in Table~\ref{tab: untrain-tcr-image-c} and Table~\ref{tab: untrain-tcr-text-c}, we observe that the untrained TCR achieves significant improvement over EATA, even without parameter update, which corroborates our observations and validates the effectiveness of the proposed TCR.

\begin{wraptable}{r}{0.45\textwidth}
\vspace{-0.4in}
    \caption{The intra-modality uniformity and inter-modality gap of different baselines after TTA under the ``Base2COCO'' setting. $\text{IU}$ and $\text{TU}$ indicate the uniformity of image and text modalities, respectively. $\text{MG}$ indicates the modality gap. 
    }
    \label{tab: ablation after TTA}
\newcommand{\tabincell}[2]{\begin{tabular}{@{}#1@{}}#2\end{tabular}}
 \begin{center}
\begin{threeparttable}
\LARGE
\resizebox{0.9\linewidth}{!}{
\begin{tabular}{l|ccc|ccc>{\columncolor{blue!8}}c}
Method & $\text{IU}$ & $\text{MG}$ & $\text{TR@1}$ & $\text{TU}$ & $\text{MG}$ & $\text{IR@1}$ \\    
\cmidrule{1-7}         
Base & 0.62 & 0.72 & 59.3 & 0.67 & 0.72 & 45.4 \\
Tent & 0.82 & 0.74 & 61.7 & 0.85 & 0.76 & 41.7 \\
EATA & 0.87 & 0.68 & 64.2 & 0.88 & 0.67 & 47.9 \\
SAR  & 0.86 & 0.70 & 63.5 & 0.74 & 0.69 & 46.6 \\
READ & 0.85 & 0.72 & 62.1 & 0.84 & 0.70 & 46.4 \\
DeYO & 0.88 & 0.68 & 65.0 & 0.86 & 0.67 & 47.3 \\
Ours & 0.93 & 0.63 & 68.9 & 0.96 & 0.64 & 48.9 \\ 
\cmidrule{1-7}   
\end{tabular}  
}  
\end{threeparttable}
	 \end{center}
\vspace{-0.3in}
\end{wraptable}
\subsection{More Ablation Results}
\label{Appendix: more ablation}
We provide more ablation studies under the ``Base2COCO'' setting to prove that TCR achieves better performance by enlarging intra-modality uniformity and rectifying the inter-modality gap.
Specifically, we present the intra-modality uniformity and inter-modality gap of different baselines after TTA in Table~\ref{tab: ablation after TTA}.
The results illustrate that i) most of the baselines improve the performance by implicitly enlarging the intra-modality uniformity and narrowing the modality gap; ii) the improvement of Tent is unstable due to the enlarged modality gap; iii) the proposed TCR achieves the highest intra-modality uniformity ($0.93$ and $0.96$) and enjoys the modality gap ($0.63$ and $0.64$) in the target domain close to that ($0.67$) in the source domain, thus contributing to boosting the performance.
Notably, we obtain the modality gap in the source domain by constructing a subset of 12,000 image-text pairs derived from the COCO, Visual Genome~\citep{VG}, CC3M~\citep{CC3M}, and SBU Captions~\citep{SBU} datasets. 
\begin{wrapfigure}{r}{0.4\textwidth}
\vspace{-0.5in}
\centering
\includegraphics[width=1.0\linewidth]{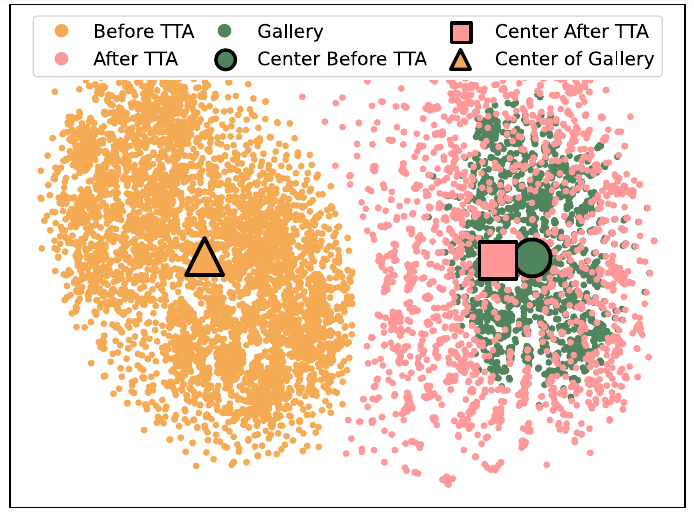}
\vspace{-0.25in}
\caption{The t-SNE visualization results of image retrieval on the query and gallery embeddings by employing the proposed TCR.}
\vspace{0in}
\label{fig: appendix_tsne}
\end{wrapfigure}
\subsection{More Visualization Result}
\label{Appendix: more visualization}
As shown in Fig.~\ref{fig: ablation}(c) of the manuscript, we have visualized the text retrieval results before/after TTA. 
Here, we provide additional visualization results of image retrieval before/after TTA under the ``Base2COCO'' setting.
The results in Fig.~\ref{fig: appendix_tsne} illustrate that samples in the query modality enjoy more scatter and the modality gap narrows after the TTA process, which proves that TCR improves performance in both TR and IR by rectifying the intra-modality uniformity and the inter-modality gap.
\end{document}